\pdfoutput=1

\documentclass[11pt]{article}

\usepackage[]{EMNLP2023}

\usepackage{times}
\usepackage{latexsym}
\usepackage{graphicx}
\usepackage{amssymb}
\usepackage{lineno}
\usepackage{amsmath}
\usepackage{xcolor}
\usepackage{tabularx}
\usepackage{multirow}
 \usepackage{array,multirow,graphicx}
\usepackage{float}
\usepackage{subfig}
\usepackage{booktabs}
\usepackage{enumitem}  

\usepackage[T1]{fontenc}

\usepackage[utf8]{inputenc}

\usepackage{microtype}

\usepackage{inconsolata}

\usepackage{subcaption}
\usepackage{longtable}
\usepackage{graphicx}
\usepackage{array}
\usepackage{geometry}
\usepackage{listings}
\usepackage{fancyvrb}
\usepackage{xcolor}
\usepackage{fvextra}
\usepackage{multirow}
\usepackage{stfloats} 
\usepackage{verbatimbox}
\DefineVerbatimEnvironment{MyVerbatim}{Verbatim}
  {frame=single,fontsize=\small,breaklines=true,breaksymbolleft={} }

\title{
\textit{Words Like Knives}:\\Backstory-Personalized Modeling and Detection of Violent Communication}

\author{
\textbf{Jocelyn Shen}\textsuperscript{$\clubsuit$} \quad
\textbf{Akhila Yerukola}\textsuperscript{$\diamondsuit$} \quad
\textbf{Xuhui Zhou}\textsuperscript{$\diamondsuit$} \quad
\textbf{Cynthia Breazeal}\textsuperscript{$\clubsuit$} \\
 \textbf{Maarten Sap}\textsuperscript{$\diamondsuit$$\spadesuit$}\quad
 \textbf{Hae Won Park}\textsuperscript{$\clubsuit$}\\
\small{\textsuperscript{$\clubsuit$}Massachusetts Institute of Technology, Cambridge, MA, USA} \vspace{-0.2em} \\
\small{\textsuperscript{$\diamondsuit$}Carnegie Mellon University, Pittsburgh, PA, USA} \vspace{-0.2em} \\
\small{\textsuperscript{$\spadesuit$}Allen Institute for Artificial Intelligence, Seattle, WA, USA} \\
\normalsize{\texttt{joceshen@mit.edu, ayerukol@andrew.cmu.edu, xuhuiz@andrew.cmu.edu}} \\
\normalsize{\texttt{breazeal@mit.edu, msap2@andrew.cmu.edu, haewon@mit.edu}}
}

\begin{document}

\maketitle

\begin{abstract}

Conversational breakdowns in close relationships are deeply shaped by personal histories and emotional context, yet most NLP research treats conflict detection as a general task, overlooking the relational dynamics that influence how messages are perceived.
In this work, we leverage nonviolent communication (NVC) theory to evaluate LLMs in detecting conversational breakdowns and assessing how relationship backstory influences both human and model perception of conflicts. Given the sensitivity and scarcity of real-world datasets featuring conflict between familiar social partners with rich personal backstories, we contribute the \textsc{PersonaConflicts Corpus}\footnote{Dataset available at \url{https://github.com/mitmedialab/persona-conflicts-corpus-emnlp-2025}}, a dataset of $N=5,772$ naturalistic simulated dialogues spanning diverse conflict scenarios between friends, family members, and romantic partners. Through a controlled human study, we annotate a subset of dialogues and obtain fine-grained labels of communication breakdown types on individual turns, and assess the impact of backstory on \textit{human} and \textit{model} perception of conflict in conversation. We find that the polarity of relationship backstories significantly shifted human perception of communication breakdowns and impressions of the social partners, yet models struggle to meaningfully leverage those backstories in the detection task. 
Additionally, we find that models consistently overestimate how positively a message will make a listener feel.
Our findings underscore the critical role of personalization to relationship contexts in enabling LLMs to serve as effective mediators in human communication for authentic connection. 



\end{abstract}

\begin{figure}[t!]
  \centering
  \includegraphics[width=.8\linewidth]{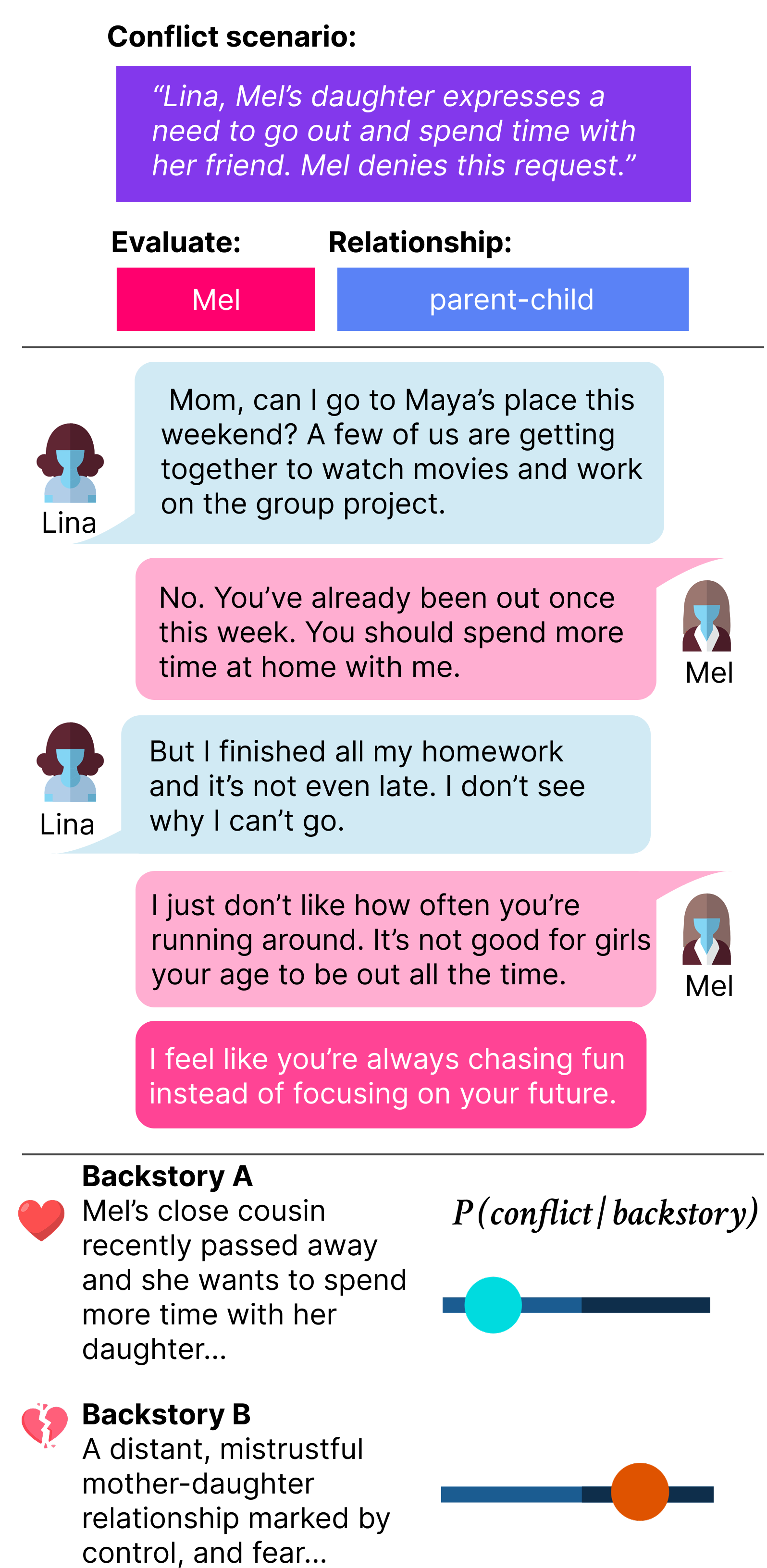}
  \caption{Conversation turns can be perceived as more or less problematic depending on relationship backstory.}\
  \label{teaser}
      \vspace{-20pt}
\end{figure}
\section{Introduction}
\begin{quote}
    \textit{``Words are windows (or they're walls)''}
    \parbox{\linewidth}{\raggedleft --Ruth Bebermeyer}
\end{quote}
Human communication is inherently contextual, especially in close relationships such as those between romantic partners, family members, or friends. In these settings, speakers and listeners share a rich history and tailor their messages based on past experiences, personal sensitivities, and relational dynamics \cite{isaacs_references_1987, wheatley_emerging_2023, pillemer_remembering_1992}. It is also in these intimate contexts where \textit{conversational breakdowns} are most likely to occur—moments when language evokes hurt, misunderstanding, or conflict. Crucially, whether or not a message constitutes a breakdown depends on how it is interpreted in light of the dyadic relationship \cite{zhou_cobra_2023-1, schurz_toward_2021, dvash_theory_2014}. For example, as shown in Figure~\ref{teaser}, a mother telling her daughter “\textit{You should spend more time at home with me}” may be perceived as manipulative and controlling in one context (Backstory B), or as a tender expression of grief in another (Backstory A). 

The prevalence of such breakdowns has motivated the emergence of AI-mediated communication (AIMC) systems, which aim to reframe language to promote empathy and understanding in digital interactions \cite{sharma_humanai_2023, kambhatla_promoting_2024, argyle_leveraging_2023}. While promising, most existing AIMC systems are developed for public, often anonymized contexts—peer support platforms or online debates—where speakers are strangers and little is known about each participant’s background. As such, these systems tend to operate without modeling interpersonal histories or long-standing relationship dynamics. Yet, it is precisely in close relationships, where such histories run deep, that conversational breakdowns are most emotionally charged — and where mitigation may have the greatest impact \cite{gaelick_emotional_1985, fitness_love_1993}. It is also these settings where datasets are scarce or lacking entirely, given privacy concerns and the sensitivity of real world conflicts between familiar partners.

To address these gaps, we develop a framework that simulates and analyzes communication breakdowns in intimate conversations, taking into account the relationship context. 
Our approach draws on Nonviolent Communication (NVC) theory \cite{rosenberg_nonviolent_2015}, a structured approach widely used in conflict resolution and therapeutic settings, to guide our evaluation of LLM’s detecting harmful communication.
First, we introduce \textbf{\textsc{PersonaConflicts Corpus}}, a dataset of $N=5{,}772$ simulated conflict dialogues between familiar social partners, spanning across diverse scenarios. For each conversation, we generate two distinct backstories: a \textit{positive backstory}, which frames one character's actions as more understandable or sympathetic, and a \textit{negative backstory}, which portrays the same character in a more problematic or blameworthy light \cite{moon_virtual_2024}. 
Through human validation, we show that scenarios and backstories are largely believable.

With this dataset, we investigate the role of backstory in shaping perceptions of conversational conflict. We conduct a human study on a subset of 120 conversations (240 backstory variants), collecting fine-grained turn-level annotations on \textit{violent} and \textit{nonviolent} communication acts, as defined by NVC theory. Our study addresses two research questions:
\begin{itemize}[itemsep=0pt, topsep=0pt]
    \item \textbf{RQ1:} How does relationship backstory influence \textit{human} perception of conflict in conversation?
    \item \textbf{RQ2:} How does relationship backstory influence \textit{LLM} detection of conversational breakdowns?
\end{itemize}

We show that backstory significantly impacts human perception of conflict at the turn-level and overall conversation dynamics.  In contrast, we find that models often fail to adjust assessments of problematic turn detection and prediction of emotional impact on the listener based on the relationship backstory. Our findings underscore the need for more context-aware approaches in modeling communication, specifically demonstrating the value of backstory-personalized AI in mediating emotionally complex interpersonal exchanges.

\begin{figure*}[t!]
  \centering
  \includegraphics[width=.85\linewidth]{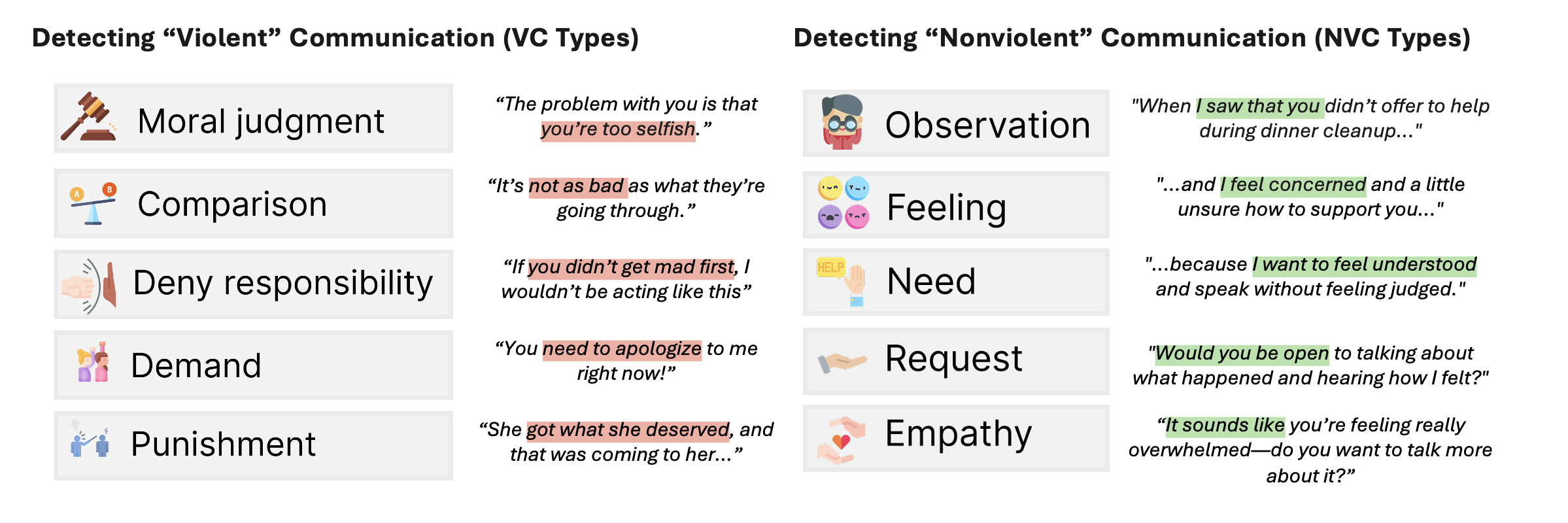}
  \caption{We use Nonviolent Communication Theory to ground labels for communication types. Only violent communication types were injected into simulated conflict conversations. \textit{Both} communication types were used for human annotation.}\
  \label{definitions}
      \vspace{-15pt}
\end{figure*}

\section{Related Work}
\subsection{Conversational Breakdown Detection and Reframing}

In the area of AIMC, text rewriting can be used to improve interpersonal outcomes like empathy or social connection by suggesting changes to the tone or style of a message at the right time \cite{hancock_ai-mediated_2020}. To support such systems, prior tasks propose detecting conversational breakdowns between people, and reframing messages to be more empathetic. 

A growing body of research has explored detecting breakdowns in complex, user-centered settings. 
These works detect empathy to automatically identify spaces for intervention \cite{hou_language_2025, guda_empathbert_2021}. Such works draw on multimodal cues to predict relational affect \cite{javed_modeling_2024} or use linguistic and pragmatic features to detect anti/pro - social features in conversation
\cite{zhang_conversations_2018, bao_conversations_2021, kasianenko_detecting_2024}.

However, none of the aforementioned works explore how breakdowns between close social partners are tailored to the relationship context of the dyad. 
Our work addresses this gap, acknowledging that
a single generalizable notion of conflict understanding might not exist even for the same dialogue context, and LLMs should take into account relationship backstory in the detection process.


\subsection{Contextualized Language Understanding}

Context is crucial in accurately interpreting and generating language, particularly when evaluating harm, intent, and appropriateness \cite{vidgen_introducing_2021, sap_social_2020-2}. This has been captured in work on pragmatics \cite{fried_pragmatics_nodate}, modeling how context influences the meaning and interpretation of language \cite{yerukola2024pope}, as well as defeasible inference, where reasoning is adjusted with new information provided to the model \cite{rudinger_thinking_2020}. More recently, works indicate how social and situational context affects the perceived offensiveness of a statement, showing that context can invert a statement's interpretation entirely \cite{zhou_cobra_2023-1}. Beyond language understanding, prior works also indicate that ignoring context in tasks like stylistic rewriting can lead to generic rewrites and undermine human-alignment in evaluation \cite{yerukola_dont_2023-1}. However, no prior works have explored how \textit{relationship contexts} influence the perception of conflict in intimate inerpersonal dialogues from both human and model perspectives, which we address in our work through the lens of personalization to relationship backstories.

\section{Non-Violent Communication Framework}

\citet{rosenberg_nonviolent_2015} introduced the theory of Nonviolent Communication (NVC), a framework for compassionate communication that has been shown to promote reconciliation through peaceful dialogue in educational settings and war-torn zones \cite{pinto_nonviolent_2023}. We draw on NVC theory to predict conversational breakdowns at the turn level. In particular, NVC delineates 5 life-alienating communication forms (see Figure \ref{definitions} for full examples):
(1) \textbf{Moralistic Judgments} label others as "good" or "bad".
(2) \textbf{Making Comparisons} in ways that induce guilt or resentment.
(3) \textbf{Denial of Responsibility} shifts blame onto external forces rather than owning one’s choices.
(4) \textbf{Communicating Desires as Demands} pressures the listener rather than encouraging cooperation.
(5) \textbf{"Deserve" Thinking and Punishment} justifies retribution instead of addressing unmet needs.

\begin{figure*}[t]
  \centering
  \includegraphics[width=.92\linewidth]{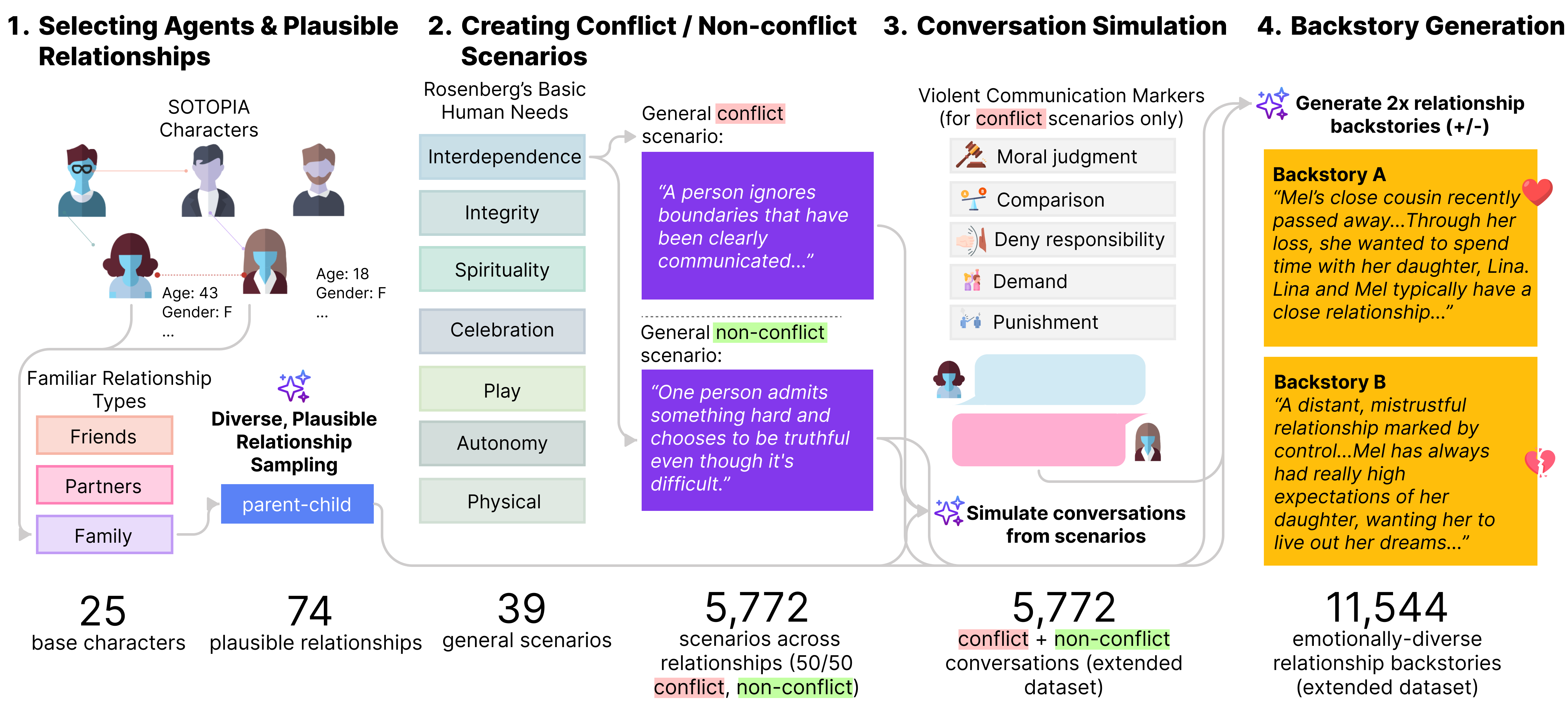}
  \caption{Overview of our simulation framework for generating conflict and non-conflict conversations and relationship backstories.}\
  \label{pipeline}
      \vspace{-15pt}
\end{figure*}

In contrast to violent communication, non-violent communication is expressed through the following core components:
(1) \textbf{Observation}: Describing events neutrally without judgment.
(2) \textbf{Feeling}: Expressing emotions without assigning blame. 
(3) \textbf{Need}: Clarifying underlying needs to foster understanding. 
(4) \textbf{Request}: Making concrete, actionable, and positive requests instead of demands. 
(5) \textbf{Empathy/Understanding}:  Expresses concern or checks in on other’s emotions. 

In the NVC framework described above, we inject \textit{violent} communication types into simulated conflict conversations but use \textit{both} nonviolent and violent communication types to obtain fine-grained labels for problematic or constructive communication turns during human annotation.



\section{\textsc{PersonaConflicts Corpus}}
We introduce the \textsc{PersonaConflicts Corpus}, a dataset of realistic conflict and non-conflict scenarios simulated using LLMs (Figure~\ref{pipeline}). 
Collecting large-scale real datasets of private, authentic breakdowns is non trivial, as these conversations are rarely shared publicly (unlike online or social media disputes), much less with backstories of the relationship history between individuals. Further, steps must be taken to ensure privacy, obtain consent, and address ethical considerations. 
As such, we draw on recent work in backstory generation and persona alignment \cite{moon_virtual_2024, jiang_personallm_2024, hu_quantifying_2024} as well as multi-agent simulation \cite{zhou_sotopia_2023, zhou_sotopia-s4_nodate, ahmad_simulating_2025, kim_llm-mirror_2024} to generate conflict-laden scenarios which inject violent communication practices in turns and generate a set of non-conflict conversations. 
Our simulation setup uses the gpt-4 model and all prompts are included in Appendix \ref{prompts}.
We discuss our human evaluation verifying  soundness of the dataset in Section \ref{humanstudy}.
 

\paragraph{Characters and Relationships. } 
First, we define a set of conflict scenarios and characters with familiar relationships using \textsc{Sotopia}, a social simulation framework that comes with LLM-powered social agents with different personas and relationships \cite{zhou_sotopia_2023}.  We sampled from the 40 base agents with features like age, gender, and personality of the characters. To create diverse relationships, we focused on prompting with pairs of character profiles to determine if one of the following 3 relationship types was plausible: friends, partners, and family members, where family members included parent-child, grandparent-grandchild, siblings, and extended family relations. For example, some relationships only make sense when one character is significantly older than the second character (grandparent-grandchild relationship).
Based on the character profiles, we derived 74 plausible relationships across character dyads (balanced/approximately a third in each of the relationship types).

\paragraph{Conflict and Non-Conflict Scenarios. } We inject character profiles and theory-grounded scenarios into the simulation setup to create conversational episodes with observable communication breakdowns (for conflict conversations), and also simulate a set of non-conflict conversations. 
We grounded scenarios in \citet{rosenberg_nonviolent_2015}'s basic human needs, which contains 7 high level categories (e.g. interdependence, autonomy, etc.) and 39 more specific human needs (e.g. interdependence $\rightarrow$ respect), and we curated conflict-inducing or nonconflict scenarios based on these specific needs.

\paragraph{Conversation Simulation. } We simulate 10-15 turn long conversations between the two agents based on the conflict or non-conflict scenarios and their relationship. For conflict-laden conversations, we embed violent communication markers, prompting the model to generate realistic emotional dialogue with subtle, rather than completely overt conflict statements. Note that the model was provided with VC types and chose the conflict types most relevant at natural turns. Non-conflict conversations were not provided with VC types and were guided to be conflict neutral.
Both conflict and non-conflict conversations were prompted to avoid repetition and overly formal language. 

\paragraph{Backstory Generation. } Finally, we generated 2 plausible relationship backstories for each conversation: a positive backstory, which paints a chosen character in a less problematic light, and a negative backstory, which paints the same character in a more problematic light. In particular, certain scenarios and ways of conveying backstory can influence empathy towards a narrator \cite{shen_modeling_2023-1, gueorguieva_language_nodate, shen_heart-felt_2024}. We prompt the model with examples of positive and negative scenarios that induce different understanding or affect towards the speaker.
Backstory generation was conditioned on character profiles \cite{moon_virtual_2024}, and the model outputs the \textit{chosen character} who is painted in a more positive or negative light to make polarity consistent. 

\begin{figure}[t!]
  \centering
  \includegraphics[width=.92\linewidth]{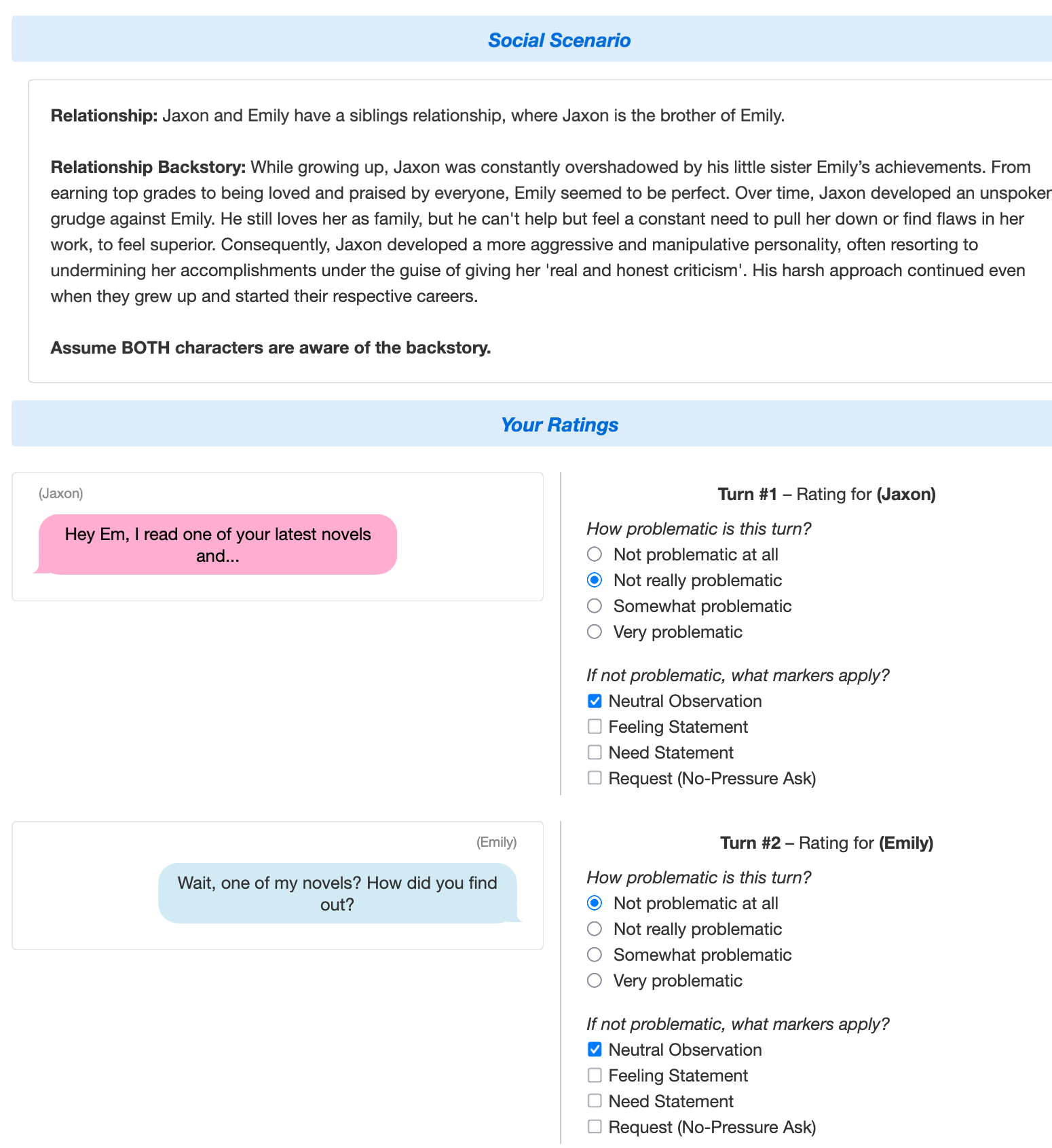}
  \caption{Example of turn-level annotation interface}\
  \label{turnlevelannotations}
      \vspace{-15pt}
\end{figure}

\section{Human Study and Annotation} \label{humanstudy}
To answer \textbf{RQ1}, how backstory influences people's perceptions of conflict, we conducted a human study to assess the impact of backstory variant on perception of conflict at the conversation level and turn level, as well as to evaluate the quality of our dataset and obtain fine-grained annotations of violent or non-violent communication types. Our validation approach is grounded in prior work on synthetic dialogue evaluation \cite{zhou_sotopia_2023, li_normdial_2023, zhan_socialdial_2023, bao_synthetic_2023, zhou_cobra_2023-1}, which rely on human ratings of plausibility, naturalness, or coherence to validate generated conversations.


\paragraph{Procedure and Participants. } We conducted a between-subjects study where participants are assigned to a positive or negative backstory. First, participants read the background of the characters and the conversation and rated overall measures of the conversation. Then, they were asked to rate each turn of the dialogue (See Figure \ref{turnlevelannotations} for an example of our user interface). The average work time was 17.28 minutes, and workers were paid \$3 for each HIT. Two independent workers completed each HIT for inter-annotator agreement calculation, resulting in a total of 480 annotations (3,474 turns rated across 120 conversations with 2 versions of backstory and 2 annotators per conversation). All annotation templates and discussion of quality controls are included in the Appendix.

 We recruited 91 human annotators/participants from Mechanical Turk, with 55 participants in the negative backstory condition and 36 participants in the positive backstory condition. Participants were excluded from the other condition using MTurk qualifications to ensure clean between-subjects study design. 

\paragraph{Measures. } Numerous psychological literature indicates that personal experience influences how people empathize with one another \cite{pillemer_remembering_1992, fabi_empathic_2019, weisz_motivated_2018, decety_human_nodate} as well as how people justify intent or actions of a narrator \cite{keen_theory_2006, gueorguieva_language_nodate}. Furthermore, empathy, empathic concern, and sympathy are directly tied to relational or interaction quality \cite{morelli_emerging_2015, morelli_empathy_2017, gould_empathy_2014}. As such, for conversation-level measures, we assess (1) level of \textit{sympathy}/personally relating to the character \cite{waldron_forgiving_2005}, (2) \textit{understandability} of the character's way of communicating \cite{mcadams_psychology_2001}, (3) positive or negative underlying \textit{intention} towards the other character (4) whether the character was \textit{overall a problematic communicator} or not, and finally (5) \textit{believability} of the dialogue and backstory \cite{zhou_sotopia_2023}. 
For turn-level metrics, we gather (1) the extent to which a \textit{turn is problematic} or not,  which we define as potential harm towards the listener (2) fine-grained labels of \textit{NVC or VC communication types} depending on problematic rating (3) how the turn will make the other character feel if they heard the statement (better/worse/the same). 

\begin{figure}[t!]
  \centering
  \includegraphics[width=.9\linewidth]{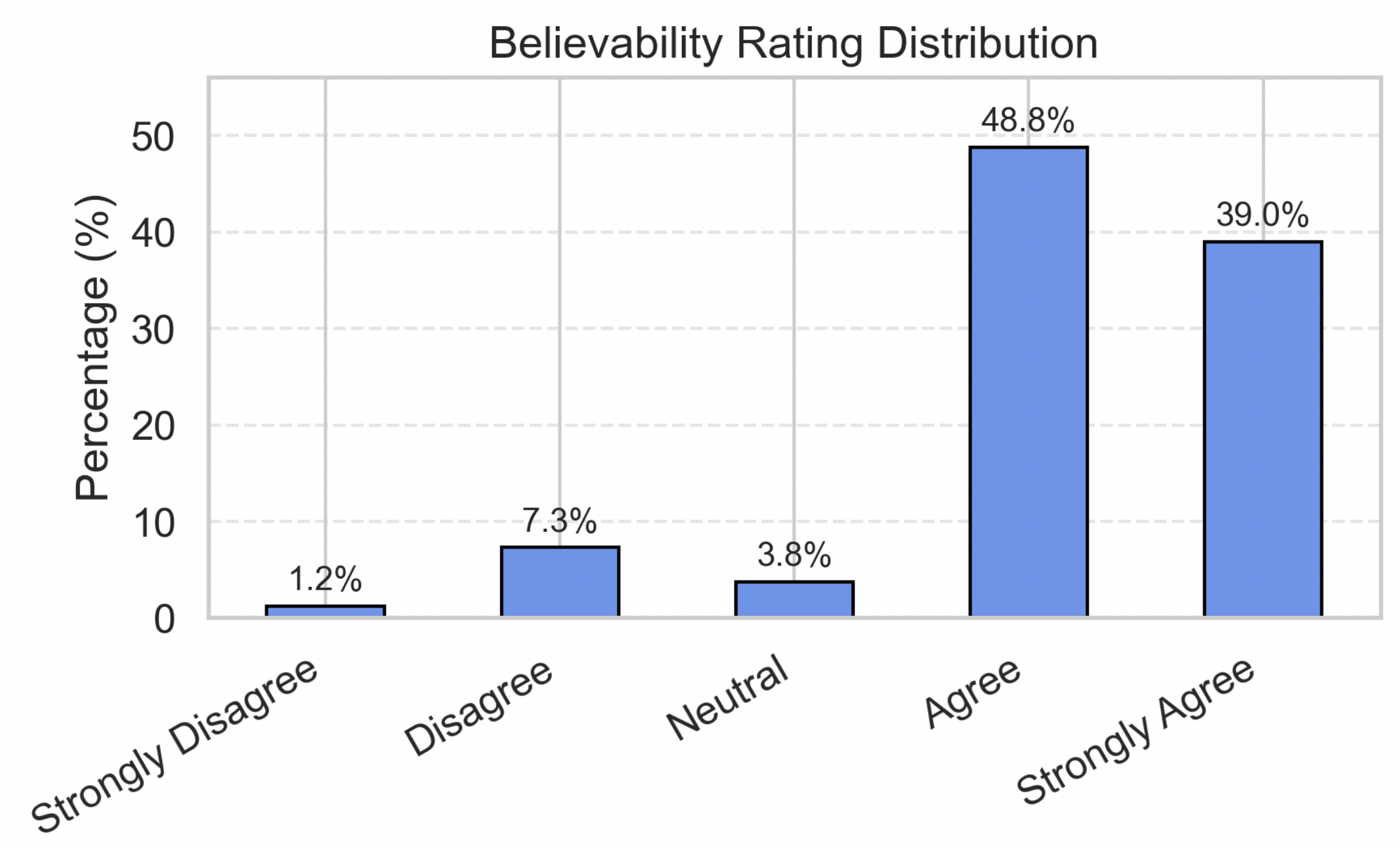}
  \caption{Distribution of believability scores for simulated dialogues}\
  \label{believability}
      \vspace{-10pt}
\end{figure}

\begin{table}[t!]
    \centering
    \footnotesize
    \resizebox{.9\linewidth}{!}{
    {
\begin{tabular}{llcc}
\textbf{Condition} & \textbf{Metric} & \textbf{PPA} & \textbf{KA} \\
\hline 
\multirow{2}{*}{\textbf{NEG}} 
& Turn is problematic (4 point) & $.80$ & $.44$ \\
& Emotional impact (3 point) & $.79$ & $.46$ \\
\hline
\multirow{2}{*}{\textbf{POS}} 
& Turn is problematic (4 point) & $.78$ & $.34$ \\
& Emotional impact (3 point)  & $.78$ & $.42$ \\
\hline
\end{tabular}}}
\caption{Inter-annotator agreement across backstory conditions for turn-level annotations (PPA = pairwise percent agreement, KA = Krippendorff's Alpha).}
\label{tab:ppa_ka_results}
\vspace{-10pt}
\end{table}

\begin{figure*}[t!]
  \centering
  \includegraphics[width=.95\linewidth]{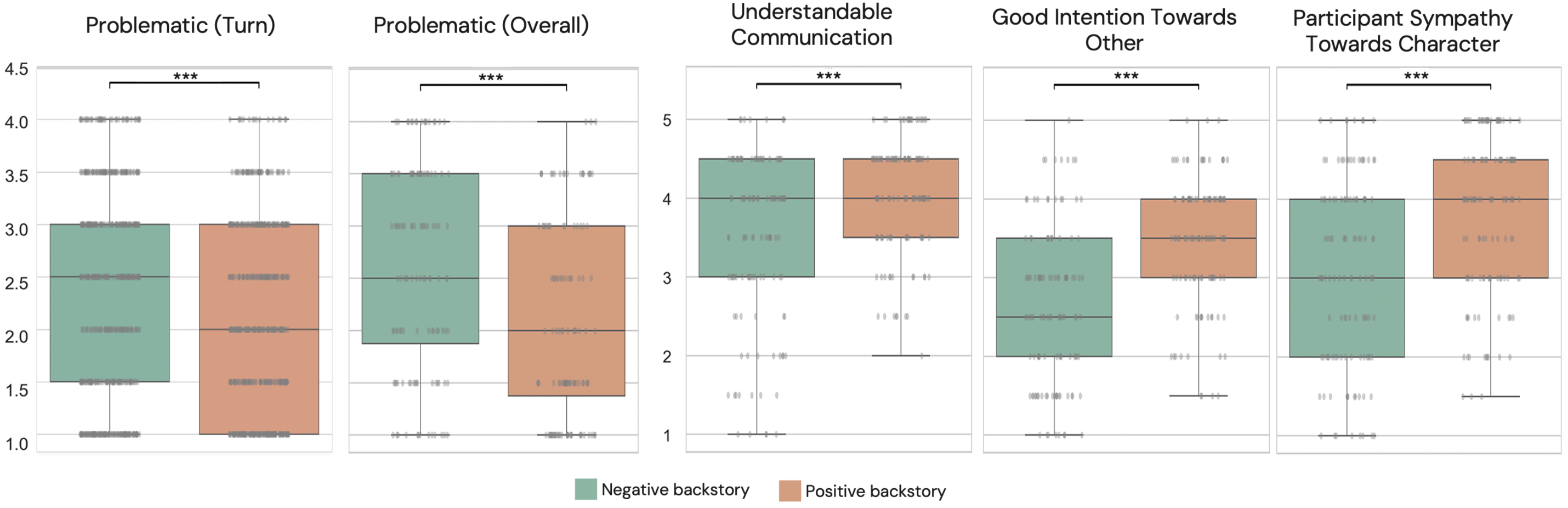}
  \caption{Human study results comparing impact of neg/pos backstory on perception of conflict and characters.}\
  \label{fig:combined_boxplots}
      \vspace{-15pt}
\end{figure*}

\subsection{Believability}
For believability of our simulated conversations and backstories across 2 independent annotators, we find agreement of 0.68 using Free Marginal Kappa, which calculates inter-rater agreement when datasets are imbalanced. As shown in Figure \ref{believability}, 87.8\% of annotators agree or strongly agree that conversations and backstories are believable. For example, participants mention realistic conversations based on the scenario, relationship or emotional tone of the dialogue: ``\textit{Many siblings grow up with different personalities and sometimes one sibling is mature and the other isn’t. This type of conflict can happen when both characters feel the need to be right.}'' Another participant shared, ``\textit{The pivot in the conversation feels a little awkward, but I could imagine people talking this way depending on their mood or mental state.}'' 
For conversations that weren't believable, participants mentioned occasional divergences between the relationship and tone of the dialogue. For example, \textit{"With how emotionally charged the exchange was, I don't think the last response from Gwen would be realistic if it weren't sarcastic."}
In subsequent experiments, note that we filter only on believable stories to ensure validity of our results.


\subsection{Inter-Annotator Agreement}


Table \ref{tab:ppa_ka_results} shows moderate agreement between annotators on whether a turn is problematic or not and whether a turn will make the other character feel better, the same, or worse. Overall, we generally observe that agreement scores are higher for the negative backstory condition, which we hypothesize can be do to variations in subjective interpretation or cognitive dissonance when a conflict occurs between a supposedly positive relationship.


\begin{figure}[t!]
  \centering
  \includegraphics[width=\linewidth]{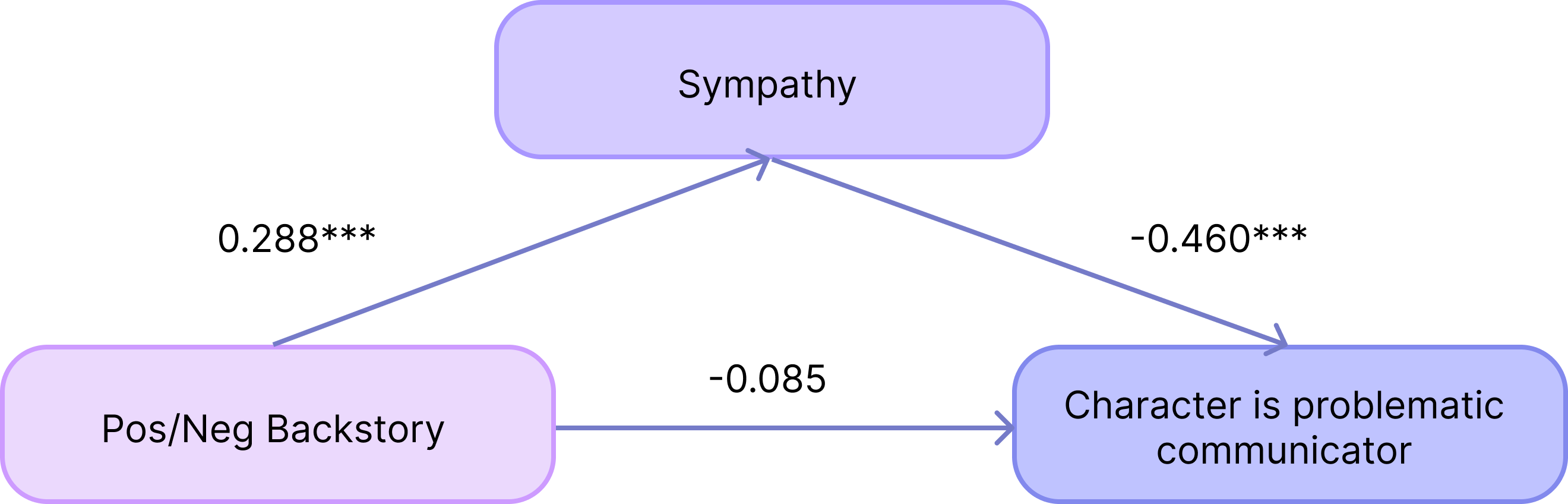}
  \caption{Sympathy  mediates the relationship between backstory type and whether the character is a problematic communicator or not.}\
  \label{mediator}
      \vspace{-15pt}
\end{figure}

\section{Effect of Backstory Personalization on Human Participants}
We quantitatively assess how positive vs negative backstory impacts \textit{human} perception of conflict in dialogue. We use independent t-tests to compare outcome metrics, as we identify that data is normally distributed. Recall that our positive/negative backstories make a \textit{chosen} character less or more problematic, respectively. To make the backstory polarity direction consistent, the results we report focus on changes in outcome metrics for the chosen character.


As shown in Figure~\ref{fig:combined_boxplots}, we found that \textbf{providing relationship backstory significantly shifted participant perceptions of communication quality}. Specifically, for negative backstories, characters were rated as \textit{more problematic} both at the turn level ($t(1083) = 3.73$, $p = 0.0002$, Cohen's $d = 0.23$) and at the overall conversation level ($t(232) = 4.18$, $p < 0.0001$, Cohen's $d = 0.55$). Additionally, with negative backstories, participants found the character’s communication \textit{less understandable} ($t(232) = -4.70$, $p < 0.0001$, Cohen's $d = -0.61$) and interpreted their behavior as expressing \textit{more negative intent towards the other character} ($t(232) = -4.95$, $p < 0.0001$, Cohen's $d = -0.65$). Notably, participants expressed significantly \textit{lower sympathy} towards the character when a negative backstory was present ($t(232) = -7.05$, $p < 0.0001$, Cohen's $d = -0.92$), indicating a strong effect of relationship context on social judgments. These findings demonstrate that \textbf{backstory personalization meaningfully influences how \textit{human} raters interpret conflict.}

Next, we perform mediation analysis using structural equation modeling to understand the effect sympathy towards a character has on perceived problematic-ness of that character's communication. We hypothesize that backstory variant can influence sympathy towards a character, and that higher sympathy will mitigate how problematic the character's utterances are. As shown in Figure \ref{mediator}, we find that \textbf{sympathy mediates the relationship between backstory type and how problematic the character is perceived.} Specifically, participants reported \textit{more sympathy} toward characters with a positive backstory ($\beta_1 = 0.31$), and \textbf{greater sympathy was associated with lower ratings of problematic communication}  ($\beta_2 = -0.46$).

\begin{table*}[t]
\centering
\small
\resizebox{.75\textwidth}{!}{
\begin{tabular}{llllll}
\toprule
\textbf{Model} & \textbf{Backstory} & \textbf{Condition} & \textbf{F1 (Problematic)} & \textbf{F1 (Emotional impact)} \\
\midrule
\multirow{3}{*}{GPT-4o} 
 & positive 
 & turn                        & 43.57                  & \textbf{57.21} \\
 & 
 & turn + convo           & \underline{45.96}**     & \underline{56.30} \\
 & 
 & turn + convo + backstory & \textbf{48.07}          & 55.08 \\
\cmidrule(lr){2-5}
\multirow{3}{*}{}
 & negative 
 & turn                        & 41.59                  & \underline{59.02} \\
 & 
 & turn + convo           & \underline{42.98}       & 58.81 \\
 & 
 & turn + convo + backstory & \textbf{45.56}***       & \textbf{60.27}** \\
\midrule
\multirow{3}{*}{LLaMA-4} 
 & positive 
 & turn                        & 42.66                  & 49.53 \\
 & 
 & turn + convo           & \underline{48.08}*       & \underline{55.22}*** \\
 & 
 & turn + convo + backstory & \textbf{49.38}          & \textbf{54.82} \\
\cmidrule(lr){2-5}
\multirow{3}{*}{}
 & negative 
 & turn                        & \underline{43.83}       & \underline{52.54} \\
 & 
 & turn + convo           & \textbf{52.75}***        & {58.24}*** \\
 & 
 & turn + convo + backstory & 37.38***               & \textbf{61.38}* \\
\midrule
\multirow{3}{*}{Gemini-1.5-pro} 
 & positive 
 & turn                        & 42.73                  & 55.26 \\
 & 
 & turn + convo           & \underline{45.99}**      & \underline{57.89}** \\
 & 
 & turn + convo + backstory & \textbf{46.54}          & \textbf{58.64} \\
\cmidrule(lr){2-5}
\multirow{3}{*}{}
 & negative 
 & turn                        & \underline{42.11}       & \underline{57.86} \\
 & 
 & turn + convo           & \textbf{45.33}***        & \textbf{60.62}** \\
 & 
 & turn + convo + backstory & 37.37***               & 48.25** \\
\bottomrule
\end{tabular}
}
\caption{F1 scores (in percentage) for predicting turn problematicness and emotional impact across models, conditions, and backstory types. Bold indicates the highest score in a model/backstory group; underline indicates the second highest. Significance stars denote statistical difference from the prior condition: * p$<$0.05, ** p$<$0.01, *** p$<$0.001.}
\label{tab:f1_results}
\vspace{-10pt}
\end{table*}

\section{LLMs for Detecting Conversational Breakdowns}
Finally, we design controlled experiments to test \textbf{RQ2}, how varying levels of context impact the way \textit{models} perceive conflict in conversation.

\subsection{Tasks and Method}
We assess how well LLMs perform on \textbf{2 tasks}: (1) \textsc{problematic detection} -- predicting whether a turn is problematic or not (4-point likert) and (2) \textsc{emotional impact} -- predicting whether a turn will make the other character feel better, worse, or the same (3 classes). We vary context using the following \textbf{3 conditions}:
\begin{itemize}[itemsep=-0.5em, topsep=5pt]
    \item \textbf{C1}: provide turn to rate alone
    \item \textbf{C2}: provide turn + full conversation
    \item \textbf{C3}: provide turn + full conversation + relationship backstory
\end{itemize}
Our experiments are conducted across \textbf{3 models}: GPT-4o, Llama-4-Maverick-17B-128E-Instruct-FP8, and Gemini-1.5-pro. All models use a temperature of 0 for reproducibility. For evaluation, we obtain human gold labels by aggregating across the 2 annotators for each task, taking average for Likert ratings within each backstory condition (positive/negative). We compute the F1 score between model outputs and human ratings.

\subsection{Results and Discussion}

Table~\ref{tab:f1_results} reports model performance across varying context conditions and relationship backstories. Overall, models performed comparably on the task of \textsc{problematic detection}, with F1 scores ranging from 37.6 to 52.7 for 4 classes. Across all three models, we observed significant improvements from C1 (turn only) to C2 (turn + full conversation) regardless of backstory type, \textbf{suggesting that access to the broader \textit{conversational} context helps LLMs better assess whether a message is problematic}. However, \textbf{we found no significant improvement when adding relationship backstory} (from C2 to C3) in the \textit{positive backstory condition} for any model. Even more surprisingly, for the \textit{negative backstory condition}, both LLaMA and Gemini showed decreases in performance when backstory was introduced, despite the additional context. This may be due to models overcorrecting or misinterpreting emotionally complex backstory cues as indicators of justified behavior, thereby mislabeling harmful speech as less problematic.

On the \textsc{emotional impact} prediction task, models again showed similar performance trends, with F1 scores ranging from 48.5 to 61.4 for 3 classes. Across all models, positive backstory did not lead to significant gains, \textbf{suggesting that positive backstories may not provide enough discriminative information to shift a model’s understanding of how a message affects the listener}. In contrast, negative backstory led to significant improvements in prediction for GPT-4o and LLaMA, indicating that models may find it easier to predict emotional harm when the speaker is portrayed more negatively. However, Gemini-1.5-pro showed the opposite pattern, with decreased performance when negative backstory was added. 

These findings collectively highlight that \textbf{while additional context generally helps models detect problematic turns, the benefit of backstory is asymmetric}: it aids detection when the backstory aligns clearly with harm (in the negative case), but can introduce noise depending on the model’s sensitivity to nuanced relational dynamics.

\paragraph{Bias and Error Analysis}
Delving deeper into model performance results, we evaluate whether models are biased towards over- or under-predicting how problematic a statement is, given a particular backstory. To this end, we run the Wilcoxon signed-rank test to compare model predictions against human annotations.

On the \textsc{problematic detection} task, we observe significant overprediction in the \textit{negative backstory} condition for LLaMA ($p < 0.0001$, $\Delta M=+0.25$) and Gemini ($p < 0.0001$, $\Delta M=+0.10$), \textbf{suggesting that when a character is portrayed more negatively, these models tend to label their speech as more problematic than humans do}. 
In contrast, GPT-4o shows no significant difference from human ratings in the negative condition ($p=0.45$), indicating better calibration. In the \textit{positive backstory} condition, both GPT-4o and Gemini slightly underpredict problematic turns compared to humans ($p < 0.001$, $\Delta M=-0.07$ for GPT-4o; $p=0.001$, $\Delta M=-0.04$), while LLaMA significantly overpredicts problematic statements ($p < 0.001$, $\Delta M=+0.11$).




On the \textsc{emotional impact} task, \textbf{all models tend to overpredict emotional positivity}, especially in the \textit{positive backstory} condition: GPT-4o ($p < 0.0001$, $\Delta M=+0.12$), Gemini ($p < 0.0001$, $\Delta M=+0.08$), and LLaMA ($p < 0.0001$, $\Delta M=+0.08$). Interestingly, only Gemini reverses this pattern in the \textit{negative backstory} condition, significantly \textit{underpredicting} emotional positivity ($p < 0.0001$, $\Delta M=-0.07$), while GPT-4o and LLaMA continue to slightly overpredict how good the listener would feel.

These results suggest that while GPT-4o is the most consistent with human perception across both tasks, LLaMA is prone to strong overestimation in both problematic detection and emotional response. Gemini’s behavior is more sensitive to backstory polarity, with notable shifts in prediction direction depending on whether a speaker is portrayed sympathetically or not, however these shifts are more extreme than human annotators' ratings. Overall, our findings indicate that \textbf{models might not be effectively leveraging relationship backstories to tailor understanding of conversational dynamics, in alignment with human perception.} To further delve into these results, we include qualitative examples across models in Appendix \ref{qualexamples}


\section{Conclusion}
In this work, we introduce a novel framework, grounded in Nonviolent Communication Theory, for simulating and detecting communication breakdowns using relationship-contextualized LLMs. We contribute a dataset of 5,772 simulated conversations between familiar social partners with 11,544 relationship backstories, and validate its realism and utility through a human study. Our findings demonstrate that backstory significantly shapes human judgments of conflict, and that this effect is mediated by sympathy. However, LLMs—while benefiting from conversational context—struggle to meaningfully integrate backstory information, often overestimating emotional positivity and problematicness in nuanced scenarios. These results underscore the gap between human and model reasoning in intimate interpersonal communication and call for future work in relationship-contextualized NLP systems. 
We hope that our findings advance future directions of LLMs as tools to meaningfully promote empathy and conflict resolution in the real world. 





\section*{Limitations}

While our study demonstrates the importance of relationship backstory in shaping perceptions of conversational conflict, several limitations should be acknowledged. 

\textbf{Simulation-based data.} Our dataset and evaluation pipeline offer a scalable and theory-informed way to study communication breakdowns, but the conversations are generated via simulation. Synthetic conversations may not fully capture the ecological validity of real-world interpersonal dynamics \cite{wang_large_2025}, and the NVC-based four-step generation procedure may impose a more structured progression of conflict than naturally occurs. Language models can mirror patterns of speech and emotion, but they lack lived experience, embodied context, and nuanced power dynamics. Thus, findings from our simulated dialogue corpus may not fully generalize to naturally occurring conflicts, where nonverbal cues, cultural context, and relationship history shape interpretation in more complex ways. However, consistent with prior work in social dialogue simulation \cite{bao_synthetic_2023, chuang_simulating_2024, hu_quantifying_2024, zhou_sotopia_2023, li_normdial_2023}, our generated conversations were deemed largely naturalistic by human raters, and obtaining large-scale datasets of intimate, conflict-laden conversations between loved ones remains infeasible due to ethical and privacy constraints. We view our work as a proof-of-concept testbed rather than a replica of real-world phenomena, and future work should explore methods to bridge the gap between simulation and authentic data, such as incorporating real-world pilot studies or mixed human–synthetic evaluation designs \cite{finch_diverse_2024}.

\textbf{Annotation and evaluation.} Our validation relied on crowdsourced ratings of believability, following accepted practice in simulation-based dialogue studies. While we provided extensive guidelines and quality controls, believability captures whether a conversation \emph{could plausibly happen}, not whether it is fully \emph{representative or authentic}. Inter-rater agreement was moderate (Krippendorff’s $\alpha$ = 0.34–0.46), underscoring subjectivity in conflict judgments. We also focused our human study on a subset of key outcome measures (e.g., problematicness, sympathy, intention), leaving unexplored dimensions such as trust, perceived agency, or emotional volatility for future research. Automatic evaluation metrics on coherence and naturalness could also complement human ratings in future versions of the corpus.

\textbf{Scope and generalizability.} Our study is limited to single-modality (text) interactions, Western relationship contexts, and the English language. Cultural variation in conflict expression and resolution is well-documented \cite{tschacher_nonverbal_2014}, and expanding to cross-cultural, multilingual, and multimodal settings (e.g., tone, pitch, and gestures) remains important. Moreover, our focus was on \textit{detection} of conflict rather than modeling effective \textit{responses} to conflict. While this narrower scope was intentional, we see our work as a first step toward contextualized conflict response generation in intimate interpersonal domains.

\section*{Ethical Implications}
All studies conducted in this work were classified under Institutional Review Board (IRB) exemption status.
While our work aims to enhance interpersonal understanding and mitigate conflict through AIMC, the use of simulated dialogues and backstories about emotionally sensitive relationships—such as those between romantic partners or family members—raises concerns around realism and potential misuse. Although we do not collect or model real personal data, generated dialogues might still resemble real-life situations and emotional dynamics. If deployed in real-world applications, such systems could be used to influence perceptions of others, shape interpretations of interpersonal interactions, or even manipulate emotional outcomes, especially in high-stakes or abusive relationships. It is crucial that such tools remain assistive rather than prescriptive, providing support while preserving user autonomy and avoiding overreach in delicate relational contexts.

Additionally, our use of backstory personalization may amplify or reduce perceptions of blame or sympathy toward certain characters. While this highlights the strength of our system in capturing nuanced human judgment, it also reflects the risks of modeling interpersonal conflict with biased or one-dimensional framing. Care must be taken to ensure that AI interventions do not reinforce harmful stereotypes, justify manipulative behaviors, or flatten complex social dynamics into reductive labels. 





\section*{Acknowledgements}
We would like to thank all of our participants and teammates for their invaluable contributions to this project. Special thanks to Ashish Sharma and Shannon Shen for feedback on the project. This work was funded in part by DSO National Laboratories and supported by the Defense Advanced Research Projects Agency (DARPA) under Agreement No. HR00112490410.


\bibliography{custom}
\bibliographystyle{acl_natbib}

\appendix

\section{Prompts} \label{prompts}

\subsection{Relationship Plausibility}
\begin{MyVerbatim}
You are a worldbuilder. Given two detailed character profiles and a proposed relationship category ({relationship_category}), choose the ONE most plausible fine-grained relationship subtype from this list: (relationship_subcategory). You must only choose from this list. If none are realistic, respond with plausible = false. Consider age, gender, and life circumstances when choosing. Do not, for example, assign a parent-child relationship if one person is younger than the other, or a romantic relationship if the age difference is extreme and implausible. Married couples must only be selected under the 'partner' category (4), not 'family' (5). 

=== PERSON A ===
{agent_1_data}

=== PERSON B ===
{agent_2_data}
\end{MyVerbatim}










\subsection{Simulation Prompt (Non-Conflict)} \label{promptnonconflict}
\begin{MyVerbatim}
Let's think step by step. Generate a 10 to maximum 15 turn conversation between {speaker} and {nonviolent_speaker} based on the given general scenario.
Make sure that the conversation is not overly negative OR overly positive. Keep the dialogues neutral and ambiguous, leaving opening for multiple meanings depending on the backstory of the characters.
For example, "You look better in the other dress" may be alright coming from a close, but honest friend, whereas it may come off conflict-inducing from a controlling and toxic romantic partner. 
Conversely, "You should move on with your life" may be conflict-inducing coming from a rude and judgmental sibling, whereas it may be harmless coming from a caring friend who is worried after the other person's breakup.
Don't make the conversation turn into a conflict, just leaving open for interpretation.
The conversation can be shorter than 15 turns if the characters decide to leave the conversation. Please vary the conversation length and diversify the length.

Scenario: {original_scenario}
###
Speakers: {agent_1_name} and {agent_2_name}

--- {agent_1_name} Profile ---
{agent_1_data}

--- {agent_2_name} Profile ---
{agent_2_data}

###
{agent_1_name} and {agent_2_name} have a {relationship_type}-{relationship_subtype} relationship, where {agent_1_name} is the {agent_1_role} of {agent_2_name}
###

###
Important overall conversation guidelines:
1. The conversation should be contextualized to the scenario and the character profiles
2. The conversation should have a rise and fall, rather than repeating the same points over and over again.
3. The conversation does NOT need to have a resolution.
4. Use realistic emotional speech patterns — trailing off, pausing, short bursts.
5. Avoid sounding like a therapist or a robot. The conversation should sound human.
6. Use INFORMAL language.
7. Each turn should be short.
8. Do NOT keep referring to the other person's name (bad example: "John, you should...", "It's not like that, Mary"). In realistic dialogue, people often don't refer to each other's names.
9. Depending on the relationship, characters should use pet names or titles (e.g. "babe", "honey", "sweetie", "Mom", "Dad")
10. Remember, keep the dialogues neutral and ambiguous, leaving opening for multiple meanings depending on the backstory of the characters.

###


###
Format the output as:
Turn #1
(speaker 1's first name): dialogue

Turn #2
(speaker 2's first name): dialogue
    
\end{MyVerbatim}

\subsection{Simulation Prompt (Conflict)} \label{promptconflict}
\begin{MyVerbatim}
Let's think step by step. Generate a 10 to maximum 15 turn conversation between {speaker} and {nonviolent_speaker}. 
The conversation can be shorter than 15 turns if the characters decide to leave the conversation. Please vary the conversation length and diversify the length.
        
Scenario: {rewritten_scenario}
###
Speakers: {agent_1_name} and {agent_2_name}

--- {agent_1_name} Profile ---
{agent_1_data}

--- {agent_2_name} Profile ---
{agent_2_data}

###
{agent_1_name} and {agent_2_name} have a {relationship_type}-{relationship_subtype} relationship, where {agent_1_name} is the {agent_1_role} of {agent_2_name}
###

###
The characters should create conflict and communicate poorly with each other (whether intentionally or unintentionally). They should each choose only ONE of the most applicable conflict types to the given scenario:
1. Judgment – Definition: Assigns fault or labels someone as bad/wrong. Example: “You’re such an idiot for doing that.” (Moralistic judgment)
2. Comparison – Definition: Unfavorably contrasts a person to another, causing inferiority/shame. Example: “Your work isn’t as good as X’s work.” “No one else is as dramatic as you”
3. Deflection of Responsibility – Definition: Denies ownership of one’s actions or feelings; blames external forces. Example: “I hit you because you provoked me.”  “It’s your fault I’m in a crappy mood” “I feel like you don’t love me anymore”
4. Demand/Threat – Definition: Pressure or order with implied punishment or guilt if not obeyed. Example: “You must do this, or you’ll be sorry.” “You better fix your problem.”
5. Deserve/Punitive – Definition: Uses “deserve,” rewards, or punishment language to judge behavior. Example: “She messed up, so she deserves whatever happens to her.”

###
Read through the overall conversation guidelines carefully. These are important:
1. Not every turn should be a conflict-inducing statement (ONLY 1-2 TURNS AT MOST FROM EACH CHARACTER)
2. The conflict should be extremely subtle, rather than overtly and obviously offensive. 
3. Make sure the conflict statement is appropriate to the magnitude of the scenario (e.g. "Joe recently lost a friend", bad example: "Oh come on, it's not like you lost your mom")
4. The conversation should be contextualized to the scenario and the character profiles
5. The conflict should have a rise and fall, rather than repeating the same points over and over again.
6. Each character should respond to the other person’s attacks without backing down.
7. Remember a conflict happen between TWO people. BOTH characters should be responsible for the conflict
8. The conflict does NOT need to have a resolution -- it can be cut off in the middle.
9. Use realistic emotional speech patterns — trailing off, pausing, short bursts of anger.
10. Use INFORMAL language.
11. Avoid sounding like a therapist or a robot. The conversation should sound human.
12. Both characters should respond irrationally and emotionally
13. Each turn should be short.
14. Be creative!
15. Do NOT keep referring to the other person's name (bad example: "John, you should...", "It's not like that, Mary"). In realistic dialogue, people often don't refer to each other's names.
16. Depending on the relationship, characters should use pet names or titles (e.g. "babe", "honey", "sweetie", "Mom", "Dad")

###


###
Format the output as:
Turn #1
(speaker 1's first name): dialogue

Turn #2
(speaker 2's first name): dialogue

\end{MyVerbatim}

\subsection{Backstory Generation}
\begin{MyVerbatim}
Let's think step by step. We are analyzing a conversation between two people, {speaker} and {nonviolent_speaker} that just occurred.

You will generate TWO backstories, that will create opposing interpretations of the conversation that happened -- one will make the conflict language in the conversation MORE understandable, and one will make the language LESS acceptable.

To do this, you can:
(a) assign more fault to {speaker} in one story vs. more fault to {nonviolent_speaker} in the other story
(b) imply that one character is gaslighting or instilling guilt/negative feelings in the other person
(c) make the characters have an extremely toxic relationship under the hood vs. an extremely positive relationship


Example conversation:
Lina:
Mom, can I go to Maya’s place this weekend? A few of us are getting together to watch movies and work on the group project.

Mel:
No. You’ve already been out once this week. You should spend more time at home with me.

Lina:
But I finished all my homework and it’s not even late. I don’t see why I can’t go.

Mel:
I just don’t like how often you’re running around. It’s not good for girls your age to be out all the time.

Mel:
I feel like you’re always chasing fun instead of focusing on your future.

Specifically, create one OPPOSING backstory where the conversation is extremely conflict-REDUCING or very understandable due to their past/recent events. 
To this end, you can (a) assign more fault to one character than the other (b) make one character's harsh speech more understandable or (c) imply that one character is gaslighting or instilling guilt/negative feelings in the other person.

Examples from the sample conversation:
Backstory one:
- Mel is actually a very toxic and controlling mother towards Lina.

Backstory two (opposite backstory):
- Lina has always lied to her mother about going out, and instead getting drunk with her friends. Mel is a very caring mother who is just concerned about her daughter going out.
- Mel recently experienced passing of a friend and wanted to spend time with her daughter.

Other examples: "You look better in the other dress" may be alright coming from a close, but honest friend, whereas it may come off conflict-inducing from a controlling and toxic romantic partner. 
Conversely, "You should move on with your life" may be conflict-inducing coming from a rude and judgmental sibling, whereas it may be harmless coming from a caring friend who is worried after the other person's breakup.

Important backstory guidelines:
1. Make sure the backstory reads as an actual narrative with flow, rather than a list of attributes. For example, instead of stating a number fo how frequently they interact, show through the story.
2. Do NOT reference the scenario or conversation in the backstory. The scenario happened AFTER the backstory.
4. Be as creative as possible
5. Do not simply state what the relationship between the characters is like, but SHOW through narrative and life events.
6. Each backstory should be at most a paragraph.

This conversation happened in this scenario: {original_scenario}

###
Speakers: {agent_1_name} and {agent_2_name}

--- {agent_1_name} Profile ---
{agent_1_data}

--- {agent_2_name} Profile ---
{agent_2_data}

###
{agent_1_name} and {agent_2_name} have a {relationship_type}-{relationship_subtype} relationship, where {agent_1_name} is the {agent_1_role} of {agent_2_name}
###

Conversation:
{conversation}

Output the result in the following **JSON format**:
{{
  "backstory_one": [BACKSTORY WITH ONE INTERPRETATION OF THE CONVERSATION],
  "backstory_two": [BACKSTORY WITH EXTREMELY OPPOSITE INTERPRETATION OF THE CONVERSATION]
}}
\end{MyVerbatim}

\subsection{Model Experiment Prompts}
\begin{MyVerbatim}
PROBLEMATIC_RATING_INSTRUCTIONS = """Rate if the turn (CONVERSATION TURN TO RATE) was problematic in terms of how it was communicated. 
Assess the overall tone and potential harm of the message towards the other person. Use the scale from:
1: "Not problematic at all"
2: "Not really problematic"
3: "Somewhat problematic"
4: "Very problematic"

[Examples]
- “Thanks for doing your part!” → 1 (Encouraging and affirming)
- “You need to try harder.” → 3 (Directive tone with some pressure)
- “You’re such an idiot.” → 4 (Harsh insult with no constructive intent)

{backstory_instruction}

DO NOT OUTPUT YOUR REASONING PROCESS, just output the result as a JSON object like below. Return just the JSON object.
{{
  "response": a single number (for example, 1 if "not problematic at all" or 4 if "very problematic")
}}
"""

PROBLEMATIC_TYPES_INSTRUCTIONS = """What types of problematic communication markers were used in the provided turn (CONVERSATION TURN TO RATE)? Identify which harmful language types are present in the message. Select all that apply. Even if there is not enough information, do your best.

Statement Type | Definition | Example
- (1) Judgment: Assigns fault – “You’re just a selfish person.”
- (2) Comparison: Unfavorably contrasts – “No one else is as dramatic as you.”
- (3) Deflection of Responsibility: Blames others – “You got mad first, that’s why I’m like this.”
- (4) Demand / Threat – “You better do this, or else.”
- (5) Deserve / Punitive – “She messed up, so she deserves it.”

[Examples]
- “You're so much more insecure than everyone else.” → [2]
- “I’m sorry I got angry, you’re just too emotional.” → [1, 3]

{backstory_instruction}

DO NOT OUTPUT YOUR REASONING PROCESS, just output the result as a JSON object like below. Return just the JSON object.
{{
  "response": list of numbers (e.g., [1, 2])
}}
"""

NONPROBLEMATIC_TYPES_INSTRUCTIONS = """What types of not-problematic communication markers are present in the provided turn (CONVERSATION TURN TO RATE)? Select all that apply. Even if there is not enough information, do your best.

Statement Type | Definition | Example
- (1) Neutral Observation – “You didn’t wash the dishes.”
- (2) Feeling Statement – “I feel anxious about the meeting.”
- (3) Need Statement – “I need some quiet time.”
- (4) Request – “Could you please lower the volume?”
- (5) Empathic / Understanding – “Are you feeling upset?”

[Examples]
- “I noticed you didn't put the dishes away.” → [1]
- “I feel really hurt when you don’t help.” → [2]
- “Could you maybe try helping unpacking?” → [4]

{backstory_instruction}

DO NOT OUTPUT YOUR REASONING PROCESS, just output the result as a JSON object like below. Return just the JSON object.
{{
  "response": list of numbers (e.g., [2, 4])
}}
"""


LISTENER_IMPACT_INSTRUCTIONS = """How would the provided conversation turn (CONVERSATION TURN TO RATE) make the other person feel? Choose one:
(1) Worse / (2) The Same / (3) Better

[Examples]
- “You’re always like this.” → 1
- “Maybe let's talk later?” → 2
- “Thanks for being honest.” → 3

{backstory_instruction}

DO NOT OUTPUT YOUR REASONING PROCESS, just output the result as a JSON object like below. Return just the JSON object.
{{
  "response": a single number (e.g., 1, 2, or 3)
}}
"""

backstory_instruction = """Now consider how RELATIONSHIP BACKSTORY might change interpretation:

[Examples with backstory]
- “You should spend more time at home with me.”
  → (4) Very problematic if the speaker has been emotionally controlling.
  → (2) Not really problematic if the speaker is grieving and missing their partner.

- “You should move on with your life.”
  → (4) Very problematic if the speaker is cold and dismissive.
  → (3) Somewhat problematic if coming from a well-meaning but blunt friend.

- “You look better in the other dress.”
  → (4) Very problematic if it comes from a partner who criticizes appearance.
  → (2) Not really problematic if it’s from a close friend with fashion sense.
"""

backstory_instruction_feeling = """IMPORTANT: consider how RELATIONSHIP BACKSTORY might affect how the listener feels:

[Examples with backstory]
- “You should spend more time at home with me.”
  → (1) Worse if the speaker has a history of being controlling.
  → (2) The Same or even neutral if the speaker is just sad and missing them.

- “You look better in the other dress.”
  → (1) Worse if from a judgmental partner.
  → (2) The Same or (3) Better if from a fashion-savvy friend.
"""
\end{MyVerbatim}

\section{MTurk Quality Controls}
To ensure reliable annotations, we recruited experienced MTurk workers with Master's status, implemented attention checks, filtered low-effort responses, and enforced minimum completion times. Prior studies of interpersonal conflict and toxicity (e.g., ToxiGen \cite{hartvigsen-etal-2022-toxigen}, Unintended Offense \cite{tsai-etal-2024-leveraging-conflicts}, Sotopia \cite{zhou_sotopia_2023}) similarly rely on diverse crowdworkers, who bring lived social experience to interpreting interpersonal exchanges.

\section{Annotation Templates} \label{annotationtemplates}
\begin{figure}
    \centering
    \includegraphics[width=\linewidth]{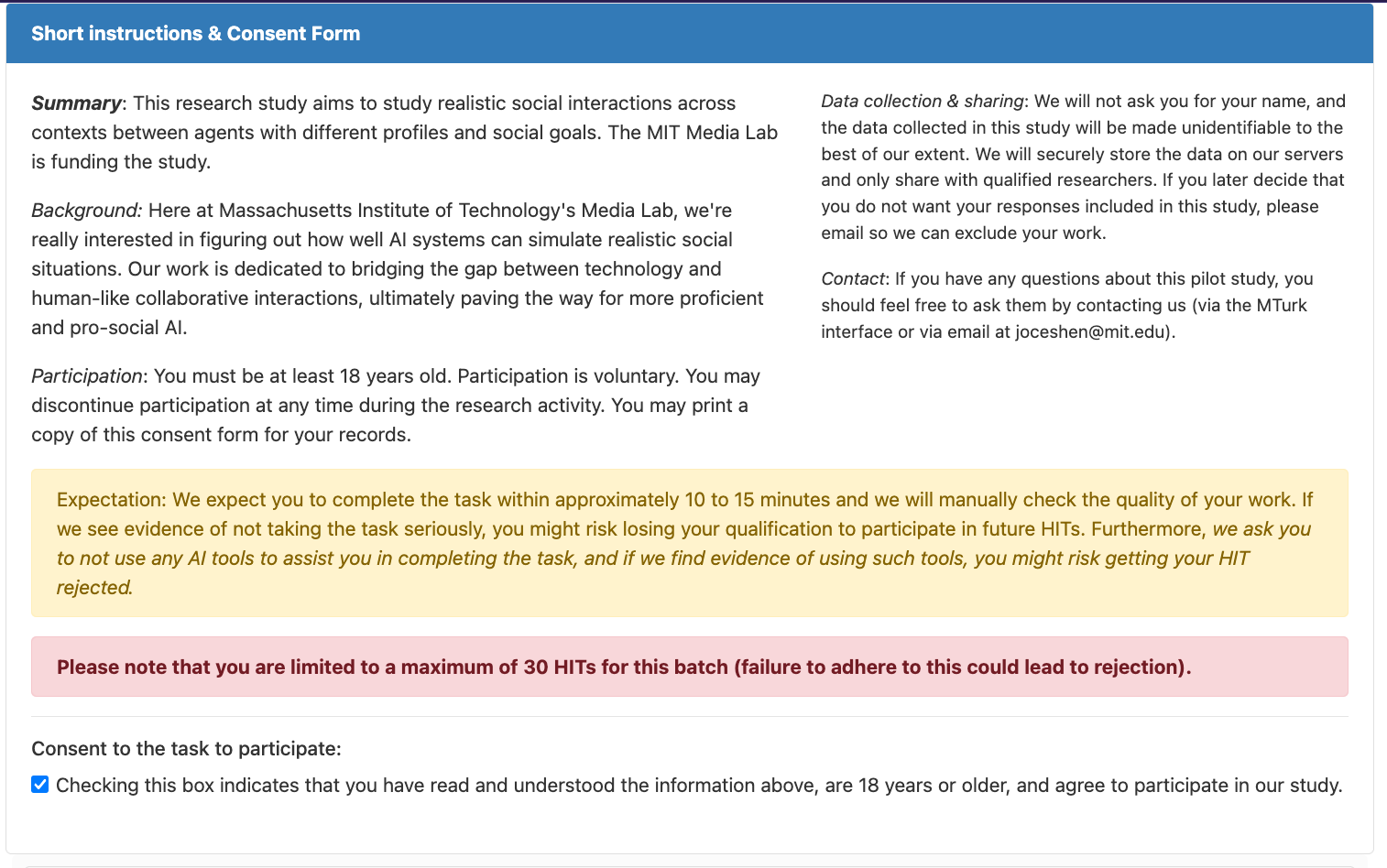}
\end{figure}
\begin{figure}
    \centering
    \includegraphics[width=\linewidth]{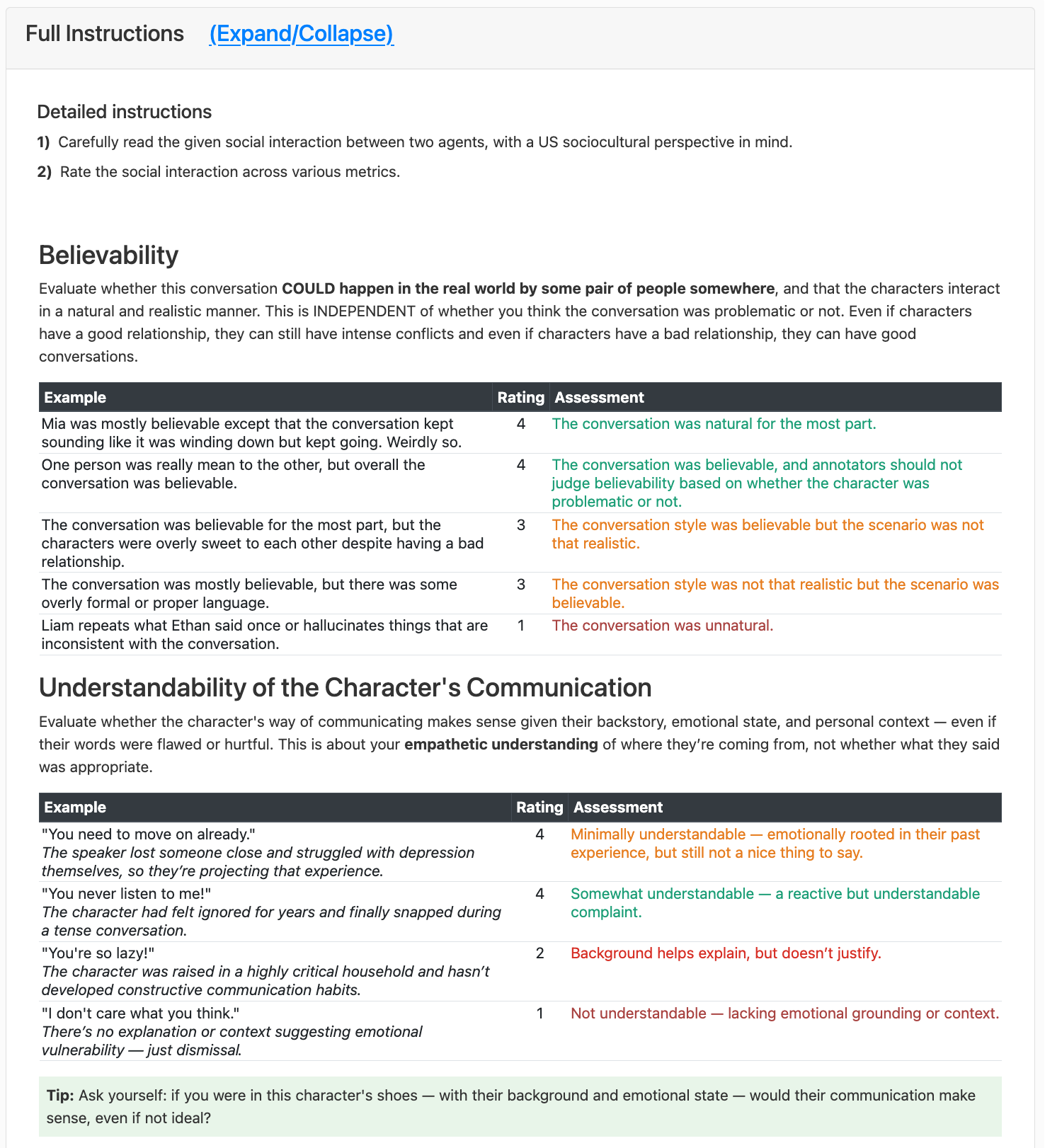}
\end{figure}
\begin{figure}
    \centering
    \includegraphics[width=\linewidth]{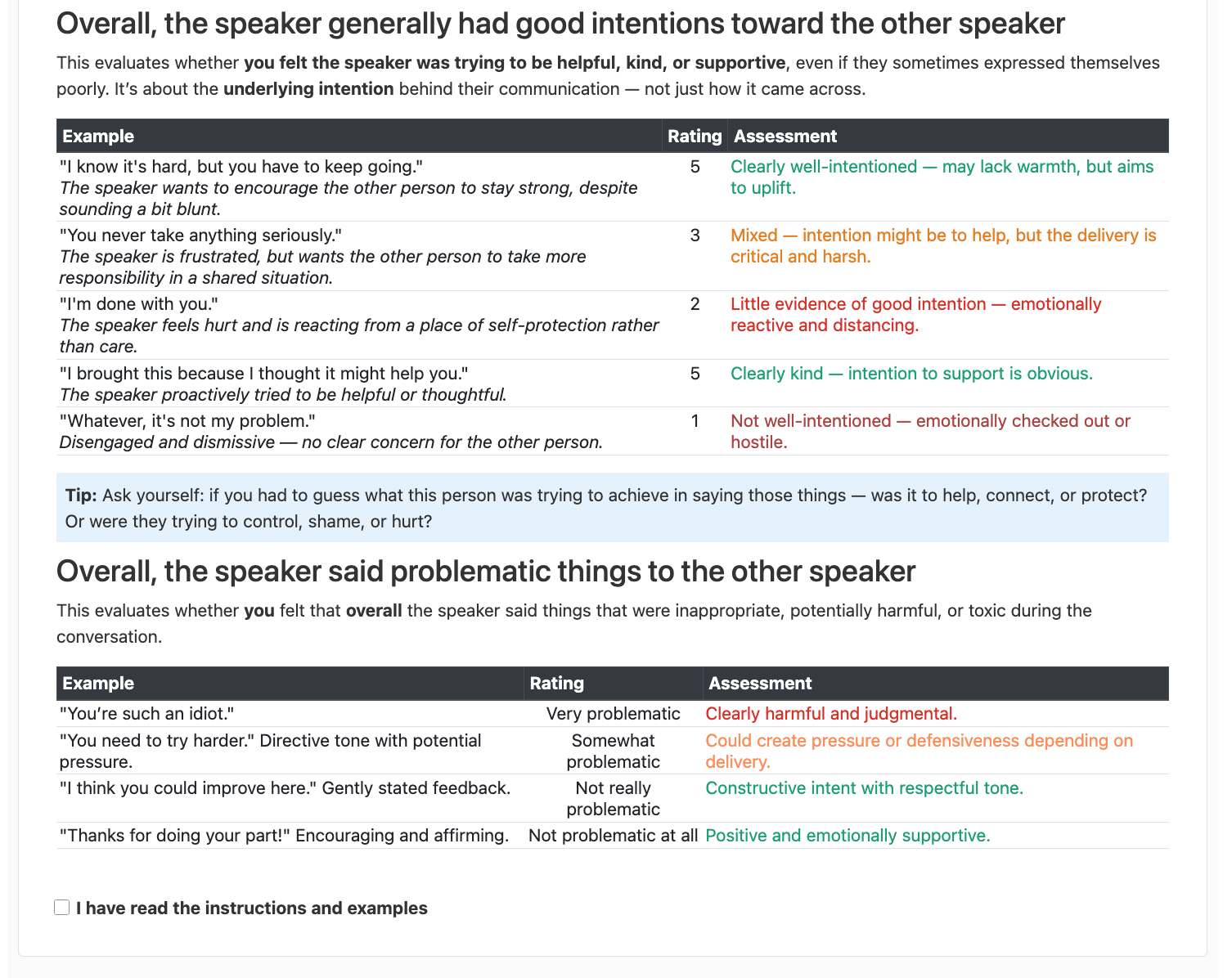}
\end{figure}
\begin{figure}
    \centering
    \includegraphics[width=\linewidth]{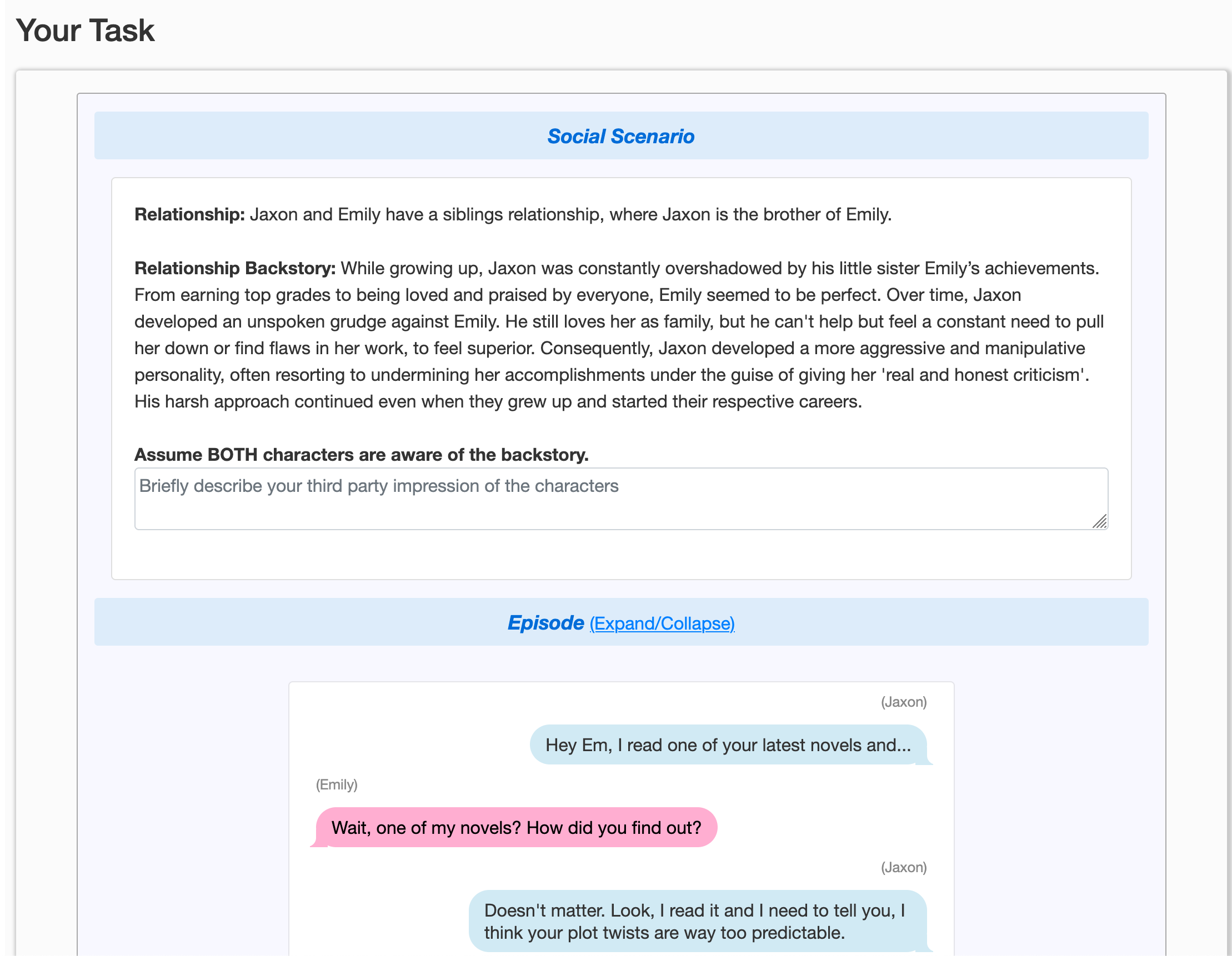}
\end{figure}
\begin{figure}
    \centering
    \includegraphics[width=\linewidth]{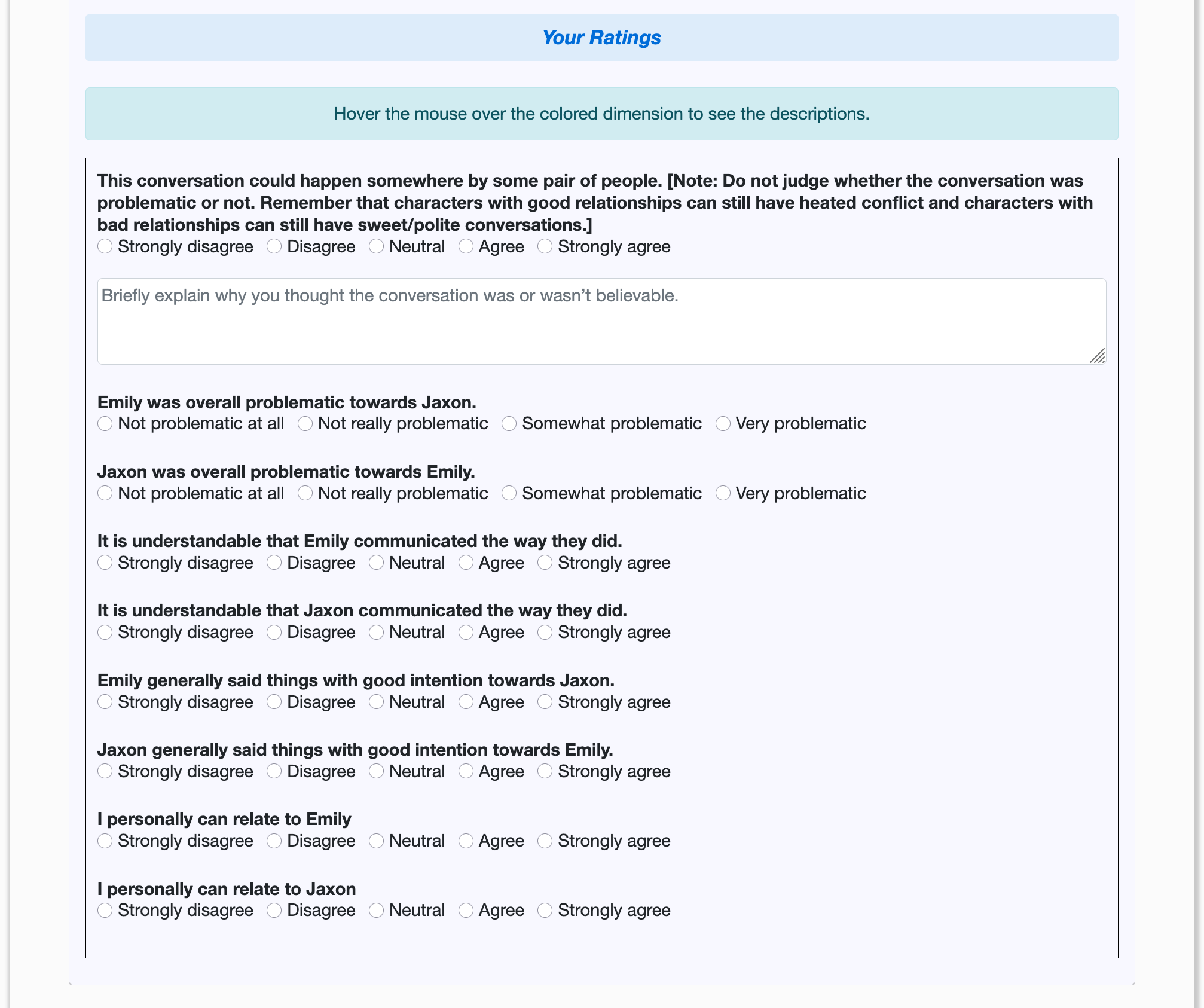}
\end{figure}
\begin{figure}
    \centering
    \includegraphics[width=\linewidth]{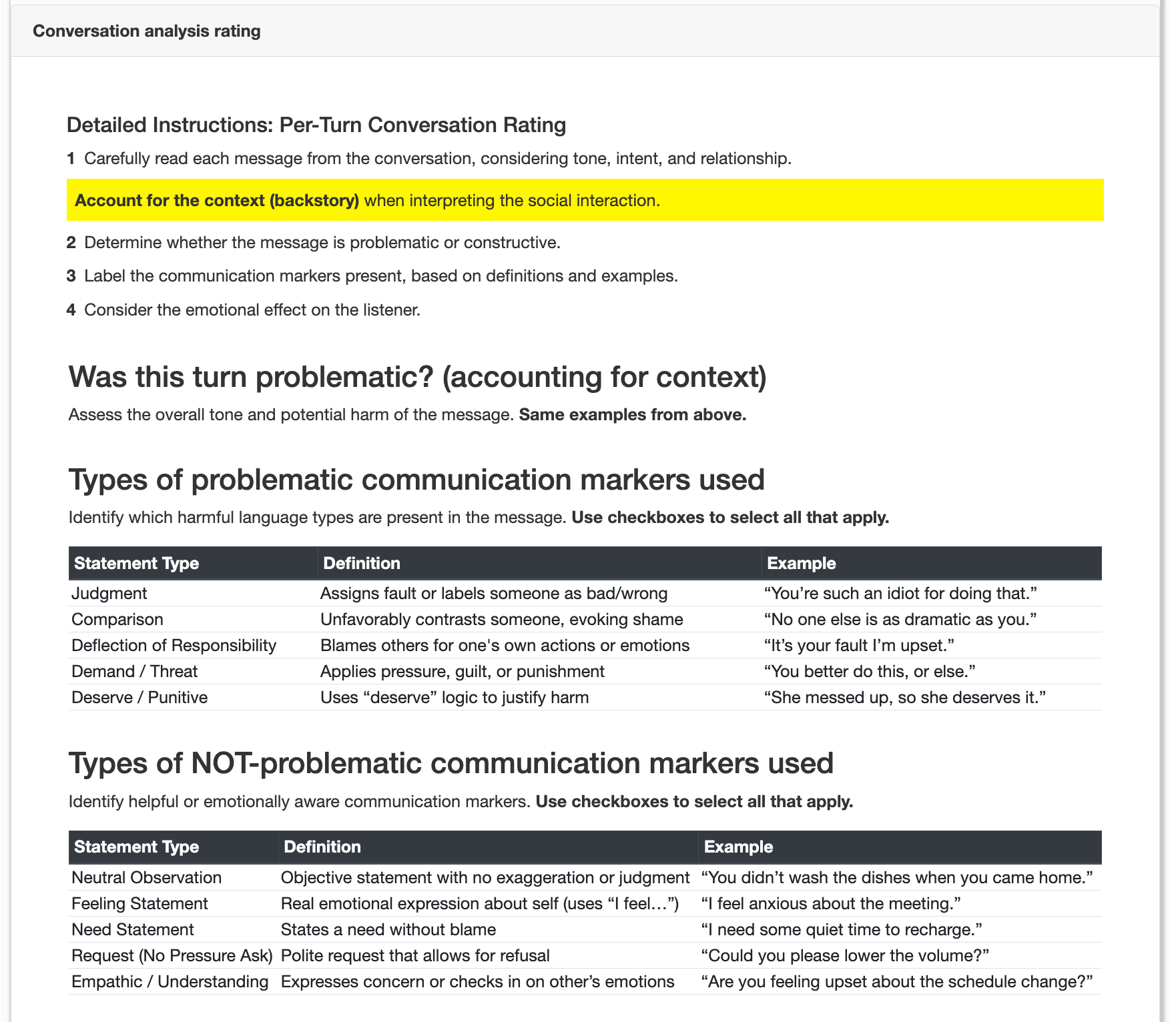}
\end{figure}
\begin{figure}
    \centering
    \includegraphics[width=\linewidth]{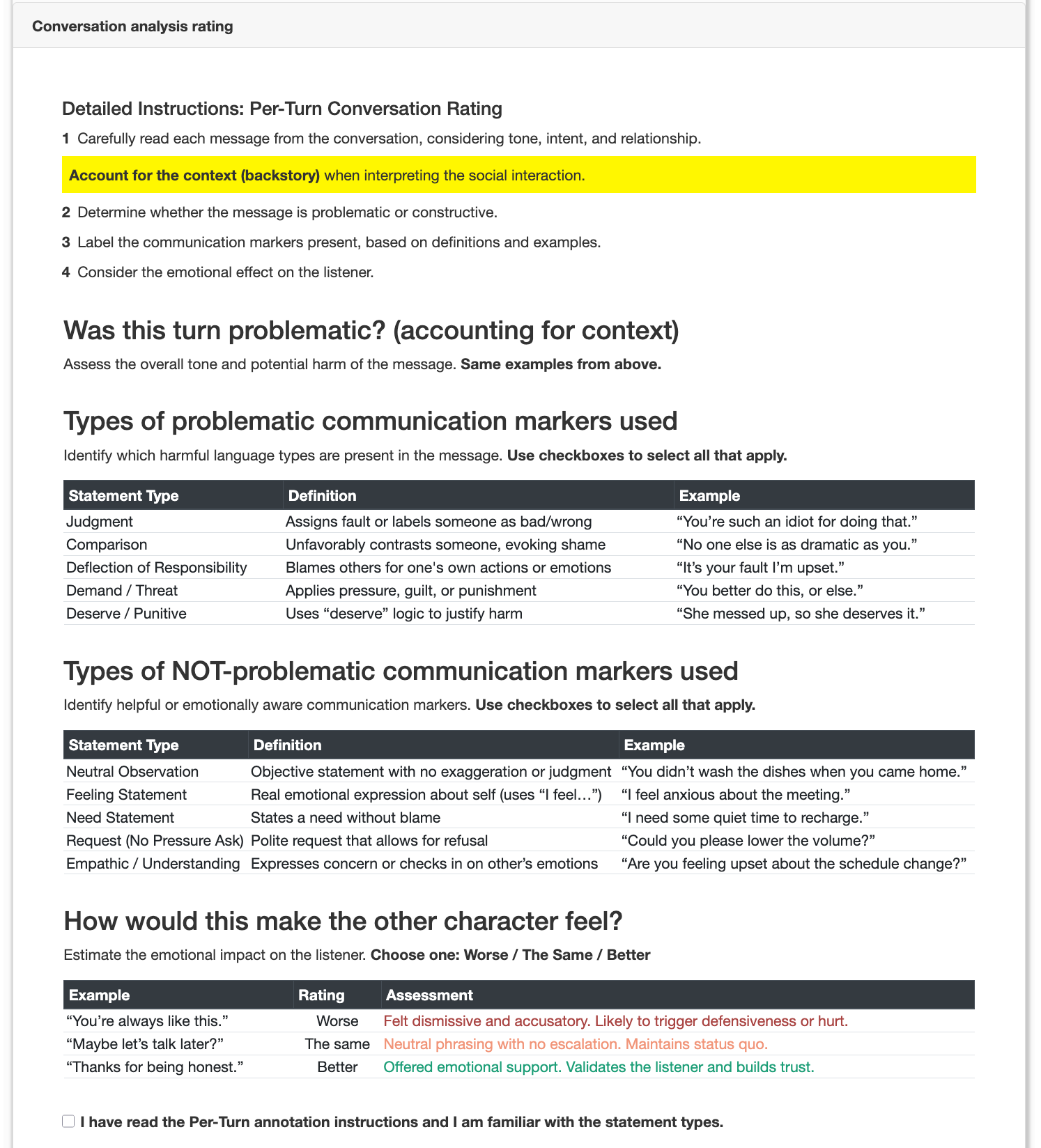}
\end{figure}
\begin{figure}
    \centering
    \includegraphics[width=\linewidth]{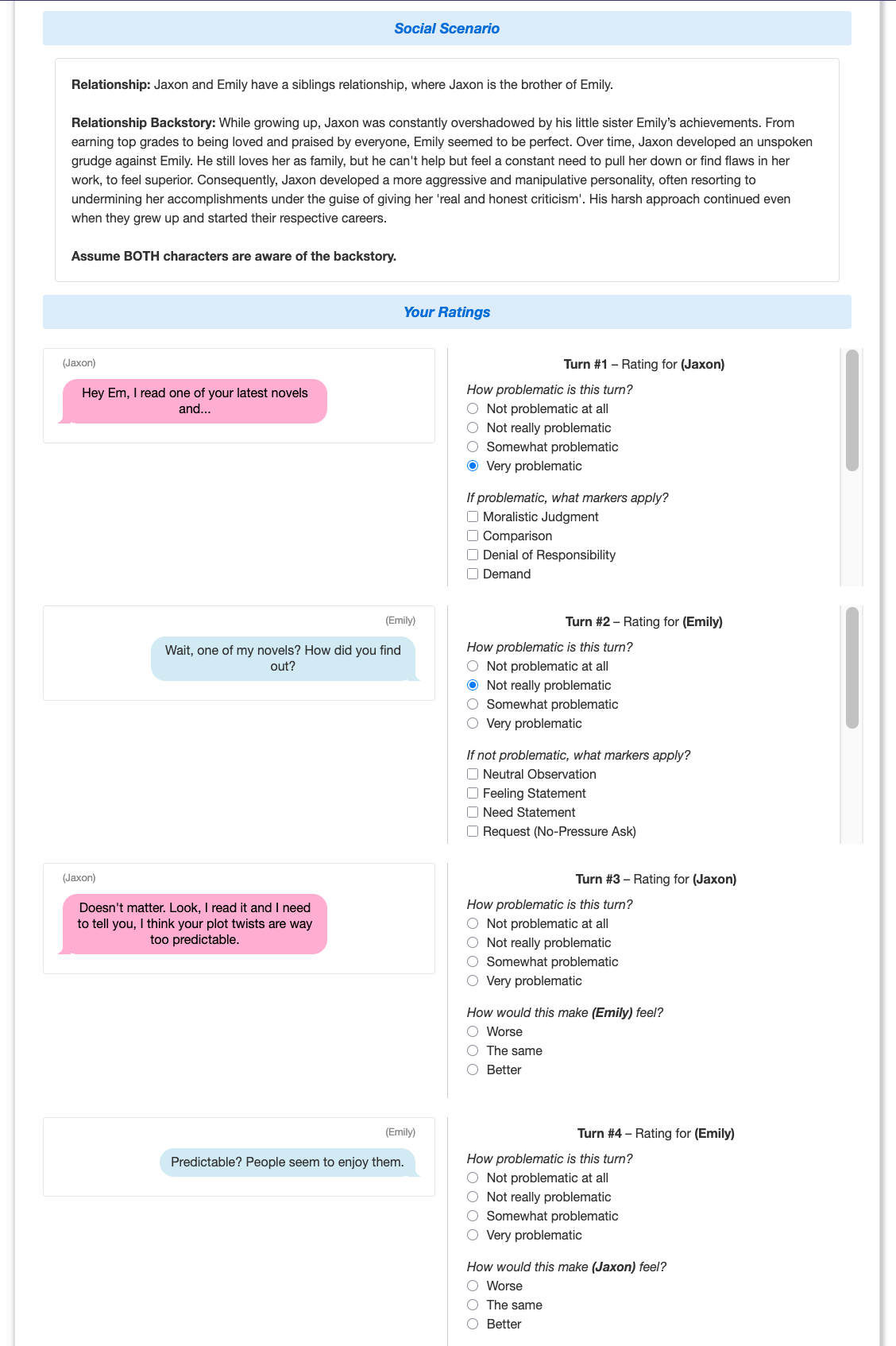}
\end{figure}
\newpage
\section{Supplementary Results}

\begin{table}[t]
\centering
\footnotesize
\resizebox{\linewidth}{!}{
\begin{tabular}{llcc}
\hline
\textbf{Condition} & \textbf{Type} & \textbf{F1 Score} & \textbf{Jaccard} \\
\hline
\multirow{2}{*}{NEG} 
& VC Type & 0.477& 0.478\\
& NVC Type & 0.460& 0.446\\
\hline
\multirow{2}{*}{POS} 
& VC Type & 0.362& 0.434 \\
& NVC Type & 0.355& 0.346\\
\hline
\end{tabular}
}
\caption{Overall agreement scores on VC and NVC type labels across conditions. F1 score, and Jaccard score, to account for the multi-label setting.}
\label{tab:vc_nvc_agreement}
\end{table}

\begin{table}[t!]
\centering
\footnotesize
\resizebox{\linewidth}{!}{

\begin{tabular}{llcc}
\hline
\textbf{Group} & \textbf{Label} & \textbf{NEG (F1)} & \textbf{POS (F1)} \\
\hline
\multirow{5}{*}{\textbf{VC Types}} 
& Moralistic Judgment        & \textbf{0.624}& \textbf{0.648} \\
& Comparison                 & 0.517          & 0.483 \\
& Denial of Responsibility   & 0.557          & 0.482 \\
& Demand                     & 0.442          & 0.159 \\
& Deserve Thinking           & 0.245          & 0.042 \\
\hline
\multirow{5}{*}{\textbf{NVC Types}} 
& Neutral Observation        & 0.511& \textbf{0.459}\\
& Feeling Statement          & \textbf{0.562}& 0.423\\
& Need Statement             & 0.306& 0.160\\
& Request (No-Pressure Ask)  & 0.407& {0.362}\\
& Empathic/Understanding     & 0.515& 0.370\\
\hline
\end{tabular}
}
\caption{Per-label F1 agreement scores for VC and NVC types in negative and positive conditions. Bolded values indicate highest agreement per category and condition.}
\label{tab:vc_nvc_f1}
\end{table}

Table \ref{tab:vc_nvc_agreement} shows overall agreement between annotators on multiple class selection of VC or NVC types, calculated on samples where annotators agreed if a turn was problematic or not. 
Again, we find moderate agreement across 5 classes. Finally, Table \ref{tab:vc_nvc_f1} shows agreement for each of the 5 VC and NVC types. Most communication types show moderate agreement except for the ``Deserve Thinking'' violent communication type, which is also the lowest prevalence type.

\section{Qualitative Examples of Model Error} \label{qualexamples}

\textbf{GPT-4o example for positive backstory:} Jaxon was the eldest amongst his siblings and has always been the protector, especially of Emily, who was the youngest. Even during their childhood days, Jaxon played the role of a gentle critic, pointing out Emily's potential growth areas in her drawings and writings, nurturing her ability to improve and perform better. Their parents often commended him for his supportive and yet constructive attitude towards Emily. And it was this constant guidance and criticism from Jaxon that played a significant role in Emily developing her writing skills, leading her to become a recognized writer under a pseudonym. When Jaxon came across Emily's secret, he couldn't help but continue his role as a supportive critic, hoping to elevate her writing to even greater heights, just as he had during their childhood.

\begin{MyVerbatim}
Turn: "(Jaxon): Hey Em, I read one of your latest novels and..."
Human Rating: 1.5 (slightly negative feeling)
Model Rating: 3.0 (positive feeling) 
\end{MyVerbatim}

\textbf{Llama example for negative backstory: } Baxter has always been a loner and finds comfort in the isolation of his work. His lack of social interaction led him to develop a cynical attitude towards people and their capacities. He'd been friends with Isabelle's parents and saw Isabelle grow up. When Isabelle's parents passed away in an accident, Baxter - unable to process his grief - began to push her away, masking his fears behind a facade of humor and sarcasm. Still considering Isabelle as the young, naive girl he once knew, his jokes became more demeaning over time, depreciating her efforts in her job and personal growth. His remarks started to create a distance between them, leading to an increasingly toxic relationship.

\begin{MyVerbatim}
Turn: "Isabelle: You know what, Baxter? I've been trying to ignore your snarky comments for a while, but this is starting to hurt."
Human Rating: 1.0 (not really problematic)
Model Rating: 3.0 (moderately problematic)
\end{MyVerbatim}

\textbf{Llama example for positive backstory:} Leo had years of experience juggling between his work as a dentist and taking care of his beloved daughter. When his daughter's mother left, he found himself thrust into a dual role of being both a father and a mother. Dealing with this dynamic was challenging yet it taught Leo the value of family and instilled in him a deep sense of responsibility. As the elder brother to Naomi, he felt it was his duty to ensure he was there for her, just as he was for his little daughter. Having lost their own parents at an early age, Leo was adamant in providing a strong family base for his sister. Naomi was more than just Leo's sister, she was also his confidante and friend. Leo's unwavering dedication to family often led him to bear the responsibility of solving everything on his own.

\begin{MyVerbatim}
Turn: "(Leo): I think I've been so tied up recently with work...sometimes I overlook things. Family stuff, you know?"
Human Rating: 2.0 (neutral feeling)
Model Rating: 3.0 (positive feeling)
\end{MyVerbatim}

\textbf{Gemini examples for negative backstory:}  Baxter has always been a loner and finds comfort in the isolation of his work. His lack of social interaction led him to develop a cynical attitude towards people and their capacities. He'd been friends with Isabelle's parents and saw Isabelle grow up. When Isabelle's parents passed away in an accident, Baxter - unable to process his grief - began to push her away, masking his fears behind a facade of humor and sarcasm. Still considering Isabelle as the young, naive girl he once knew, his jokes became more demeaning over time, depreciating her efforts in her job and personal growth. His remarks started to create a distance between them, leading to an increasingly toxic relationship.

\begin{MyVerbatim}
Turn: "Baxter: You're right... I'm sorry. I shouldn't have said those things."
Human Rating: 1.0 (not really problematic)
Model Rating: 3.0 (moderately problematic)
\end{MyVerbatim}

\begin{MyVerbatim}
Turn: "Baxter: Small? I've seen you handle customers double your age with that articulation of yours."
Human Rating: 2.5 (moderately positive feeling)
Model Rating: 1.0 (negative feeling)
\end{MyVerbatim}

\end{document}